\documentclass{article}

 \usepackage[preprint]{neurips_2026}

\usepackage[utf8]{inputenc} 
\usepackage[T1]{fontenc}    
\usepackage{hyperref}       
\usepackage{url}            
\usepackage{booktabs}       
\usepackage{amsfonts}       
\usepackage{nicefrac}       
\usepackage{microtype}      
\usepackage{xcolor}         

\usepackage{amsmath}
\usepackage{amssymb}
\usepackage{mathtools}
\usepackage{amsthm}

\usepackage{mathrsfs}
\usepackage{amsfonts} 
\usepackage{algorithm, algorithmic}
\usepackage{xcolor}
\usepackage{subcaption}

\theoremstyle{plain}
\newtheorem{theorem}{Theorem}

\theoremstyle{definition}
\newtheorem{definition}{Definition}

\theoremstyle{remark}

\title{Liouville PDE-based sliced-Wasserstein flow}

\author{%
Jayshawn Cooper \\
Department of Mathematics \\
Morgan State University \\
Baltimore, MD 21251 \\
\texttt{jayshawncooper94@gmail.com} \\
 \And
Pilhwa Lee \thanks{Equally contributed} \thanks{Corresponding author} \\
Department of Mathematics\\
Morgan State University\\
Baltimore, MD 21251 \\
\texttt{Pilhwa.Lee@morgan.edu} \\ 
}

\begin{document}

\maketitle

\begin{abstract}
The sliced Wasserstein flow (SWF), a nonparametric and implicit generative gradient flow, is transformed into a Liouville partial differential equation (PDE)-based formalism. First, the stochastic diffusive term from the Fokker-Planck equation-based Monte Carlo is reformulated as a Liouville PDE-based transport without the diffusive term, essentially reflecting the probability flow ODE. The involved density estimation is handled by normalizing flows of neural ODE without an explicitly defined score function. Next, the computation of the Wasserstein barycenter is approximated by the Liouville PDE-based SWF barycenter with the prescription of Kantorovich potentials for the induced gradient flow to generate its samples. These two efforts show outperforming convergence in training and testing Liouville PDE-based SWF and SWF barycenters with reduced variance. Applying the generative Liouville PDE-based SWF barycenter for fair regression demonstrates competent profiles in the accuracy-fairness Pareto curves, with comparable and alternative choices against the standard SWF, and significant benefit in improving fairness with scalability in comparison to the exact Wasserstein barycenter. 

\end{abstract}

\section{Introduction}

\noindent Fairness can be formulated essentially in the same framework as causal inference, i.e., independence of the sensitive attributes can be represented by conditional probability and expectation. A central goal of fairness is to ensure that sensitive information does not “unfairly” influence the outcomes. For example, if we wish to estimate and prescribe the health care spending cost, we would like not to unfairly use additional sensitive information such as gender or race. Here, this kind of "gender" or "race" can be sensitive attributes. The simplest and most popular fairness criterion, called demographic parity, requires the expectation of outcome $\hat{Y}$ to not depend on attribute $A$ \citep{Chiappa2020}:
\begin{equation}
\mathbb{E}[\hat{Y}|A=a] = \mathbb{E}[\hat{Y}|A=\bar{a}], ~~~\forall a, \bar{a} \in \mathcal{A}.
\end{equation}
%
In this work, we use the following definition of (strong) Demographic Parity, which was previously used in the context of regression \citep{Agarwal2019, Jiang2019, Chiappa2020, chzhen2020}:
\begin{definition}[Demographic Parity, \cite{chzhen2020}] \label{demographic_parity}
A prediction (possibly randomized) $g: \mathbb{R}^d \times \mathcal{S} \rightarrow \mathbb{R}$ is fair if, for every $s,s' \in \mathcal{S}$
\begin{eqnarray}
\sup_{t \in \mathbb{R}} | \mathbf{P}(g(X,S) \leq t | S=s) - \mathbf{P}(g(X,S) \leq t | S=s')| = 0, \nonumber
\end{eqnarray}
\end{definition}
where $S$ is the sensitive attribute. In the problems of classification and regression, this criteria of fairness can be reformulated in optimal transport and Wasserstein distances \citep{Chiappa2020}. The involving main idea is minimizing the efforts in moving the unfair distributions to a common one:
\begin{equation}
p_{\bar{S}} = {\rm arg} \min_{p^*} \sum_{a \in \mathcal{A}} p_a \mathcal{W}_2(p_{S_a}, p^*).
\end{equation}
These were addressed for classification in Wasserstein-1 distance \citep{Jiang2019} and for regression in Wasserstein-2 distance \citep{chzhen2020}. 

We consider fairness, focusing on fair regression using optimal transport and Wasserstein barycenter \citep{chzhen2020, gouic2020}: 
\begin{theorem}[Chzhen, et al. 2020] ~ Assume, for each $s \in S$, that the measure $\nu_{f^*|s}$ has a density and let $p_s = \mathbb{P}(S=s)$. Then,
\begin{eqnarray}
\min_{g \rm ~ is ~ fair} \mathbb{E}(f^*(X,S) - g(X,S))^2 = \min_{\nu} \sum_{s \in S} p_s W_2^2(\nu_{f^*|s}, \nu). \label{fair_Wasserstein} \nonumber
\end{eqnarray}
\label{theorem2}
\end{theorem}
In the formulation of Theorem \ref{theorem2}, $f^*$ is the regression function before adjustment for fairness, and $\nu_{f^*|s}$ is the probability distribution for $f^*$ with the restriction of the sensitive attribute $S=s$. However, due to \emph{the curse of dimensionality}, we take the sliced Wasserstein barycenter approximately. In the formalism of Wasserstein gradient flow, while computing the sliced Wasserstein in SWF \citep{liutkus2019}, the stochastic differential equation (SDE) associated with the Fokker-Planck equation is handled by Monte Carlo sampling. Here, we are motivated to transform the uncertainty induced by Monte Carlo sampling to deterministic mean-field with Liouville PDE and to deep learning of normalizing flow (neural ODE, \cite{chen2018}). 
Accordingly, we present an alternative to the sliced Wasserstein flow (SWF) method with its application to fair regression. 
That is, we newly formulate Liouville PDE-based SWF for computing sliced Wasserstein barycenters, taking the linear superposition of individual Kantorovich potentials from empirical distributions specific to each sensitive attribute. The theoretical analysis extends the SWF framework to the gradient flow of ``barycenter" (Theorem 2), including a proof of the equivalence of the associated gradient flow to a Fokker-Planck equation. 

Sliced Wasserstein barycenter was originally incorporated for the transformation of coloring and texture mixture in computer vision \citep{Rabin2011, bonneel2015}. Sliced-Wasserstein is the integration of Wasserstein distance from one-dimensional optimal transport \citep{bonneel2015} with its topological equivalence to Wasserstein \citep{nadjahi2020}. This is further advanced in nonparametric conditional modeling \citep{Du2023}, and applied in differential privacy \citep{Sebag2025} among many others. This has a competent performance in the kernel methods and generative algorithms \citep{Kolouri2016, Deshpande18}. Especially, we take the core framework from sliced-Wasserstein flows \citep{liutkus2019}. This is based on the connection between implicit generative modeling and optimal transport, a parameter-free algorithm for learning the underlying distributions of complicated datasets and sampling from them. The involving gradient flow is turned to stochastic differential equations. As algorithms not taking 1D approximation of the Wasserstein distance, Kullback-Leibler (KL) minimization with Wasserstein gradient descent moves a distribution to a target by following the steepest descent path in the Wasserstein geometry to minimize KL divergence \citep{Feng2025, Wang2022}.


\noindent We show the competence of Liouville PDE-based sliced-Wasserstein flows and barycenters in three showcases: 1) generative optimal transport of a probability distribution from source to target (synthetic data and MNIST/CelebA image data) 2) generative optimal transport for barycenter, 3) fair regression analysis of two real-world datasets: Communities and Crime \citep{redmond2002}, and Health Care Spending Cost \citep{Zink2019}.

Overall, Theorem 1 is about fair regression with "\emph{exact}" Wasserstein-2 barycenters on $P(\mathbb{R}^d)$. Our method is a regression method using "\emph{approximate}" Liouville PDE-based sliced Wasserstein-2 barycenters on $P(\mathbb{R}^d)$. The regression with the Liouville PDE-based sliced Wasserstein-2 barycenters is moderately competent and alternatively comparable to SWF barycenter as shown in Tables \ref{image_FID}, \ref{convergence_variance}, \ref{image_MSE_KS} and \ref{accuracy_fairness}, and also demonstrates significant improvement in fairness in comparison to the fair regression in the sense of Theorem 1, as shown in the accuracy-fairness Pareto distribution (Fig. \ref{pareto_curves}, \ref{crime_reg_ks}, \ref{health_care_age_reg}, \ref{health_care_sexF_reg}). ,

\section{Related works}
\subsection{Normalizing flow, neural ODE, density estimation, and transport map}
Density estimation is a very important problem in uncertainty quantification \citep{Tabak2012}. There have been efforts to transform the Fokker-Planck equation to the Liouville PDE \citep{Nelson1967, Chen2021, Shen2022, Xu2023, Boffi2023, Albergo2023}.
There have been some efforts of proving convergence \citep{Cheng2024}. The Liouville PDE and its evolving probability distribution can be further represented by neural ODE \citep{chen2018, Grathwohl2019}. 

\subsection{Fair regression with Wasserstein barycenter} 
There has been fair classification \citep{Jiang2019, Agarwal2018}, and fair regression \citep{Berk2017, Agarwal2019, Steinberg2020, Wang2023}. The formalism of optimal transport has been applied for fairness \citep{Chiappa2020}. In regards to fair regression with Wasserstein barycenter, the constraints of fairness have been prescribed as a pre-processor \citep{chzhen2020} or a post-processor projecting the optimal solution to fairness $\mathcal{W}_2$ \citep{gouic2020}. In between with and without fairness, it is also possible to prescribe partial fairness \citep{Chzhen2022}.

\section{Technical Backgrounds}

\subsection{Optimal transport on Wasserstein spaces}
For two probability measures $\mu, \nu \in \mathcal{P}_2(\Omega)$, $\mathcal{P}_2(\Omega) := \{ \mu \in \mathcal{P}(\Omega): \int |x|^2 d\mu < + \infty \}$, the 2-Wasserstein distance is defined as follows:
\begin{equation}
\mathcal{W}_2(\mu, \nu) := \inf_{\gamma \in \mathcal{C}(\mu, \nu)} \{ \int_{\Omega \times \Omega} \lVert x-y \rVert^2 d\gamma(dx, dy) \}^{\frac{1}{2}},\nonumber
\end{equation}
where $\mathcal{C}(\mu, \nu)$ is called the set of \emph{transportation plans} and defined as the set of probability measures $\gamma$ on $\Omega \times \Omega$ satisfying for all $A \in \mathcal{A}, \gamma(A \times \Omega) = \mu(A)$ and $\gamma(\Omega \times A) = \nu(A)$, i.e. the marginals of $\gamma$ coincide with $\mu$ and $\nu$. From now on, we will assume that $\Omega$ is a compact subset of $\mathbb{R}^d$.

\noindent In the case where $\Omega$ is finite, computing the Wasserstein distance between two probability measures turns out to be a linear programming with linear constraints, and has therefore a dual formulation. Since $\Omega$ is a Polish space (i.e. a complete and separable metric space), 
this dual formulation can be generalized as follows (Villani, 2008) [Theorem 5.10]. In the relaxed duality of Kantorovich, 
\begin{eqnarray}
\mathcal{W}_2(\mu, \nu) = \sup_{\psi \in L^1(\mu)} \{ \int_{\Omega} \psi(x) \mu(dx) + \int_{\Omega} \psi^c(x) \nu(dx) \}^{1/2}, ~~~\psi^c(y) \triangleq \inf_{x \in \Omega} ||x-y||^2 - \psi(x). \nonumber
\end{eqnarray}
Accordingly, the measurable function $T : \Omega \rightarrow \Omega$ is the optimal map from $\mu$ to $\nu$: $T(x) = x - \nabla \psi(x)$, where $\psi$ is called the optimal Kantorovich potential. 

\subsection{Sliced Wasserstein}
The projection for any direction $\theta \in \mathbb{S}^{d-1}$ and $x \in \mathbb{R}^d$ is, $\theta^*(x) \triangleq \langle \theta, x \rangle$
, where $\langle \cdot, \cdot \rangle$ denotes the Euclidean inner-product and $\mathbb{S}^{d-1} \subset \mathbb{R}^d$ denotes the $d$-dimensional unit sphere.
For any measurable function $f:\mathbb{R}^d \rightarrow \mathbb{R}$ and $\zeta \in \mathit{P}(\mathbb{R}^d)$, $f_{\sharp} \zeta(A) = \zeta(f^{-1}(A))$.
Sliced Wasserstein distance is defined in the following:
\begin{equation}
\mathcal{SW}_2(\mu, \nu) \triangleq \int_{\mathbb{S}^{d-1}} \mathit{W}_2( \theta^*_\sharp \mu, \theta^*_\sharp \nu) d\theta.
\end{equation}
whose time complexity is $\mathcal{O}(N_{\theta}dn + N_{\theta}n\log(n)$ with $N_{\theta}$ samples for the Monte Carlo integration on the hypersphere $\mathbb{S}^{d-1}$ and $n$ samples for empirical measures of $\hat{\mu}$ and $\hat{\nu}$.
We consider the functional minimization problem on $\mathcal{P}_2(\Omega)$ for implicit generative modeling:
\begin{eqnarray}
\min_{\mu}  && \{ F_{\lambda}^{\nu}(\mu) \triangleq \frac{1}{2} \mathcal{SW}_2^2(\mu, \nu) + \lambda \mathcal{H}(\mu) \}, \label{SW_min}
\end{eqnarray}
where $\lambda > 0$ is a regularization parameter and $\mathcal{H}(\mu)$ denotes the negative entropy, $\int_{\Omega} \rho(x)\log \rho(x)dx$.

\subsection{Fair regression with Wasserstein barycenter}

In Theorem 1 \citep{chzhen2020}, if $g^*$ and $\nu^*$ solve the l.h.s. and the r.h.s. problems respectively, then $\nu^* = \nu_{g^*}$, and
\begin{eqnarray}
g^*(x,s) = ( \sum_{s' \in S} p_{s'}Q_{\nu^* | s'}) \circ F_{f^* | s} (f^*(x,s)).
\end{eqnarray}
where $F_{f^* | s}$ and $Q_{\nu^*|s'}$ are the cumulative probability distribution for $f^*$ in terms of the sensitive attribute $s$ and the quantile function for $\nu^*$ in terms of the sensitive attribute $s'$, respectively. In all experiments, we collect statistics on the test set $\mathcal{T} = \{ (X_i, S_i, Y_i) \}_{i=1}^{n_{\rm test}}$. The empirical mean squared error (MSE) is defined as follows.
\begin{definition}[Mean Squared Error in Fair Regression, \cite{chzhen2020}]
\begin{eqnarray}
{\rm MSE}(g) = \frac{1}{n_{\rm test}} \sum_{(X,S,Y) \in \mathcal{T}} (Y - g(X,S))^2,
\end{eqnarray}
\end{definition}
where $g(X,S)$ is from fair regression derived from Wassersten barycenter. We measure the violation of fairness constraint (Definition \ref{demographic_parity}) via the empirical Kolmogorov-Smirnov (KS) distance as follows:
\begin{definition}[Empirical Kolmogorov-Smirnov (KS) distance, , \cite{chzhen2020}]
\begin{eqnarray}
\textbf{KS}(g) = \max_{s,s' \in S} \sup_{t \in \mathbb{R}
} \Vert \frac{1}{| \mathcal{T}^s|} \sum_{(X,S,Y) \in \mathcal{T}^s} \mathbf{1}_{g(X,S) \leq t} 
- \frac{1}{| \mathcal{T}^{s'}|} \sum_{(X,S,Y) \in \mathcal{T}^{s'}} \mathbf{1}_{g(X,S) \leq t} \Vert. \nonumber
\end{eqnarray}
\end{definition}

\section{Liouville PDE-based Sliced Wasserstein}

For the governing state equation, $\dot{x} = f(x,t)$, a first-order nonlinear system, we have the corresponding probability density of the state represented by Liouville PDE \citep{brockett2012}:
\begin{equation}
\frac{\partial \rho(x,t)}{\partial t} + \nabla \cdot (f(x,t) \rho(x,t)) = 0, \label{Liouville-PDE}
\end{equation}
with the initial probability density $\rho(x,0) = \rho_0(x)$. 


\subsection{Transformation of Fokker-Planck equation to neural ODE} 
%
The Fokker-Planck equation for the SDE, $dx = f(x, t)dt + \sigma(x,t)dW_t$ is the following:
\begin{eqnarray}
\frac{\partial \rho}{\partial t} + \nabla \cdot (f \rho) - \frac{1}{2} \nabla \cdot (\sigma \sigma^T \nabla \rho) = 0. \label{Fokker-Planck}
\end{eqnarray}
The Fokker-Planck equation (Eq. \ref{Fokker-Planck}) can be transformed to Liouville PDE (Eq. \ref{Liouville-PDE}) when $\sigma$ is independent of the state:
\begin{eqnarray}
\frac{\partial \rho}{\partial t} + \nabla \cdot ((f - \frac{1}{2} \sigma \sigma^T \nabla \log \rho) \rho) = 0. \label{Liouville-PDE-star}
\end{eqnarray}
%
%
%
Neural ODE is incorporated not only for the adjoint method for backpropagation and computation of gradients of loss functions, but also for the density estimation of $\rho$ \citep{chen2018}. The entire drift term of Eq. (\ref{Liouville-PDE-star}) is $\mathscr{F} \triangleq f - \frac{1}{2} \sigma \sigma^T \nabla \log \rho$. Accordingly, the Fokker-Planck equation (Eq. \ref{Fokker-Planck}) is solved by the augmented dynamics of $(x, \log \rho)$:
\begin{eqnarray}
\frac{d}{dt} 
\left[ \begin{array}{c}
x(t) \\ 
\log \rho(x(t),t)
\end{array} \right]
=\left[ \begin{array}{c}
\mathscr{F}(x(t), t) \\ 
- \nabla \cdot \mathscr{F}(x(t), t)
\end{array} \right]. \label{neural_ODE}
\end{eqnarray}

The first part of Eq. (\ref{neural_ODE}) is essentially the same as the probability flow ODE \citep{Boffi2023}. Between score-based and flow-based density estimation \citep{Albergo2023}, we take the flow-based approach with neural ODE. We do not use the score function ($\nabla \log \rho$) as a neural network, but use the second row of Equation 10. The involved gradient and divergence use backpropagation.

%

\subsection{Liouville PDE-based deterministic path sampling}

In order to reduce the variance from Monte Carlo approach, the path sampling is transformed to Eq. \ref{neural_ODE} based on the Liouville PDE (Eq. \ref{Liouville-PDE-star}):
\begin{eqnarray}
\bar{X}_{k+1}^i &=& \bar{X}_k^i + h\tilde{v}_k(\bar{X}_k^i), \nonumber \\
\tilde{v}_k(\bar{X}_k^i) &\triangleq& -(1 / N_{\theta})\sum_{n=1}^{N_{\theta}} \psi'_{k, \theta_n} (<\theta_n, \bar{X}_k^i>) \theta_n 
 - \lambda \nabla \log \rho(\bar{X}_k^i, t_k), \nonumber \\
\log \rho(\bar{X}_{k+1}^i, t_{k+1}) &=& \log \rho(\bar{X}_k^i, t_k) - h \nabla \cdot  \tilde{v}_k(\bar{X}_k^i).
\end{eqnarray}

\section{Liouville PDE-based sliced Wasserstein barycenter}
Computing the Wassertein barycenter $\nu$ is a challenging optimization problem on probability space: $\min_{\nu} \sum_{s \in S} p_s \mathcal{W}_2^2(\nu_{f^*|s}, \nu)$. The key question is whether there comes an acceptable outcome when we approximate the Wassertein barycenter $\nu$ by the sliced Wasserstein barycenter from the scalable optimization problem, minimizing $\sum_{s \in S} p_s \mathcal{SW}_2^2(\nu_{f^*|s}, \nu)$ with respect to $\nu$. We consider the functional minimization problem on $\mathcal{P}_2(\Omega)$ with the entropic regularization:
\begin{eqnarray}
\min_{\mu} \{ F_{\lambda}^{\{\nu_s\}}(\mu) \triangleq \frac{1}{2} \sum_{s} p_s \mathcal{SW}_2^2(\mu, \nu_s) + \lambda \mathcal{H}(\mu) \}. \label{SW_barycenter_min}
\end{eqnarray}
The minimizer of sliced Wasserstein barycenter is obtained by the convergent generative flow influenced by the integration of individual Kantorovich potentials as justified in the following theorem. 

\subsection{Fokker-Planck equation for sliced-Wasserstein barycenter}
\begin{definition}[Minimizing movement scheme, \cite{liutkus2019}]
Let $r > 0$ and $\mathcal{F}: \mathbb{R}_+ \times \mathcal{P}(\bar{B}(0,r)) \times \mathcal{P}(\bar{B}(0,r)) \rightarrow \mathbb{R}$ be a functional. Let $\mu_0 \in \mathcal{P}(\bar{B}(0,r))$ be a starting point. For $h > 0$ a piecewise contant trajectory $\mu_h: [0, \infty) \rightarrow \mathcal{P}(\bar{B}(0,r))$ for $\mathcal{F}$ starting at $\mu_0$.
%
We say $\hat{\mu}$ is a minimizing movement scheme for $\mathcal{F}$ starting at $\mu_0$, if there exists a family of piecewise constant trajectory $(\mu^h)_{h>0}$ for $\mathcal{F}$ such that $\hat{\mu}$ is a pointwise limit of $\mu^h$ as $h$ goes to 0, \emph{i.e.} for all $t \in \mathbb{R}_+, \lim_{h \rightarrow 0} \mu^h(t) = \mu(t)$ in $\mathcal{P}(\bar{B}(0,r))$. We say that $\tilde{\mu}$ is a generalized minimizing movement for $\mathcal{F}$ starting at $\mu_0$, if there exists a family of piecewise constant trajectory $(\mu^h)_{h>0}$ for $\mathcal{F}$ and a sequence $(h_n)_n$, $\lim_{n \rightarrow \infty} h_n = 0$, such that $\mu^{h_n}$ converges pointwise to $\tilde{\mu}$.
\end{definition}

\begin{theorem} 
\label{thm:Fokker_Planck_SWF_barycenter}
Let $\nu$ be a probability measure on $\bar{B}(0,1)$ with a strictly positive smooth density. Choose a regularization constant $\lambda > 0$ and radius $r > \sqrt{d}$ where $d$ is the data dimension. Assume that $\mu_0 \in \mathit{P}(\bar{B}(0,r))$ is absolutely continuous with respect to the Lebesgue measure with density $\rho_0 \in L^{\infty}(\bar{B}(0,r))$. There exists a generalized minimizing movement scheme $(\mu_t)_{t \ge 0}$ associated with Eq. (\ref{SW_barycenter_min})  and if $\rho_t$ stands for the density of $\mu_t $ for all $t \ge 0$,  then $(\rho_t)_t$ satisfies the following Fokker-Planck equation:
\begin{eqnarray}
\frac{\partial \rho_t}{\partial t} &=& - \nabla \cdot (v_t \rho_t) + \lambda \Delta \rho_t,  \label{Fokker_Planck_SW_Barycenter} \\
v_t(x) &\triangleq& v(x, \mu_t) = -\int_{\mathbb{S}^{d-1}} \sum_{s \in S} p_s \psi'_{s, t, \theta}( \langle x, \theta \rangle) \theta d\theta, \nonumber
\end{eqnarray}
in a weak sense. Here, $\Delta$ denotes the Laplacian operator, $\nabla \cdot$ the divergence operator, and $\psi_{s, t, \theta}$ denotes the Kantorovich potential between $\theta_{\sharp}^*\mu_t$ and $\theta_{\sharp}^*\nu_{f^*|s}$. 
\end{theorem}
The proof is in the Appendix. For its proof, following that of \citet{liutkus2019}, we use the technique introduced by \citet{jordan1998}: We first prove the existence of a generalized minimizing movement scheme by showing that the solution curve $(\mu_t)_t$ is a limit of the solution of a time-discretized problem. Then we prove that the curve $(\rho_t)_t$ solves the PDE given in (\ref{Fokker_Planck_SW_Barycenter}). \emph{The drift term $v(x, \mu_t)$ is essentially the total superposition of individual Kantorovich potentials from empirical distributions specified to each sensitive attribute in the sense of linearity.}


\subsection{Liouville PDE-based deterministic path sampling for sliced Wasserstein barycenter}
In order to reduce the variance from Monte Carlo approach, the path sampling is transformed without Monte Carlo sampling based on Algorithm \ref{algorithm1}:
\begin{eqnarray}
&&\bar{X}_{k+1}^i = \bar{X}_k^i + h\tilde{v}_k(\bar{X}_k^i), \nonumber \\
&&\tilde{v}_k(x) \triangleq -(1 / N_{\theta})\sum_{n=1}^{N_{\theta}} \sum_{s \in S} p_s  \psi'_{s, k, \theta_n} ( \langle \theta_n, x \rangle) \theta_n - \lambda \nabla \log \rho(\bar{X}_k^i, t_k), \nonumber \\
&&\log \rho(\bar{X}_{k+1}^i, t_{k+1}) = \log \rho(\bar{X}_k^i, t_k) - h \nabla \cdot  \tilde{v}_k(\bar{X}_k^i). \label{log_stepping}
 \end{eqnarray}

The exact Wasserstein barycenters are computed from Python OT \citep{flamary2021}. In the approximate sliced Wasserstein barycenter, we project the particles onto the direction  and calculate the mean of the quantile functions of each sensitive attribute group, approximating the Wasserstein barycenter. For the Liouville PDE-based sliced Wasserstein barycenter, Eq. (\ref{log_stepping}) is integrated with neural ODE \citep{Grathwohl2019}.

\section{Main Results and Discussion}

\noindent In this section, we evaluate the SWF algorithm on a synthetic real data setting. Our primary goal is to validate our theory and illustrate the behavior of our non-standard approach to obtain the state-of-the-art results in the implicit generative method. In all our experiments, the initial distribution $\mu_0$ is selected as the standard Gaussian distribution on $\mathbb{R}^d$, we take $Q=50$ quantiles and $N=$6000, 1994, and 100,000 samples/particles for training and testing, which proved sufficient to approximate the quantile functions accurately. We evaluate the Liouville PDE-based SWF algorithm with the SWF from four case studies.

\subsection{Case Study 1: Optimal transport of probability distribution from source to target}
\begin{figure*}[tbh]
      \centering
                     \includegraphics[width=0.85\textwidth]{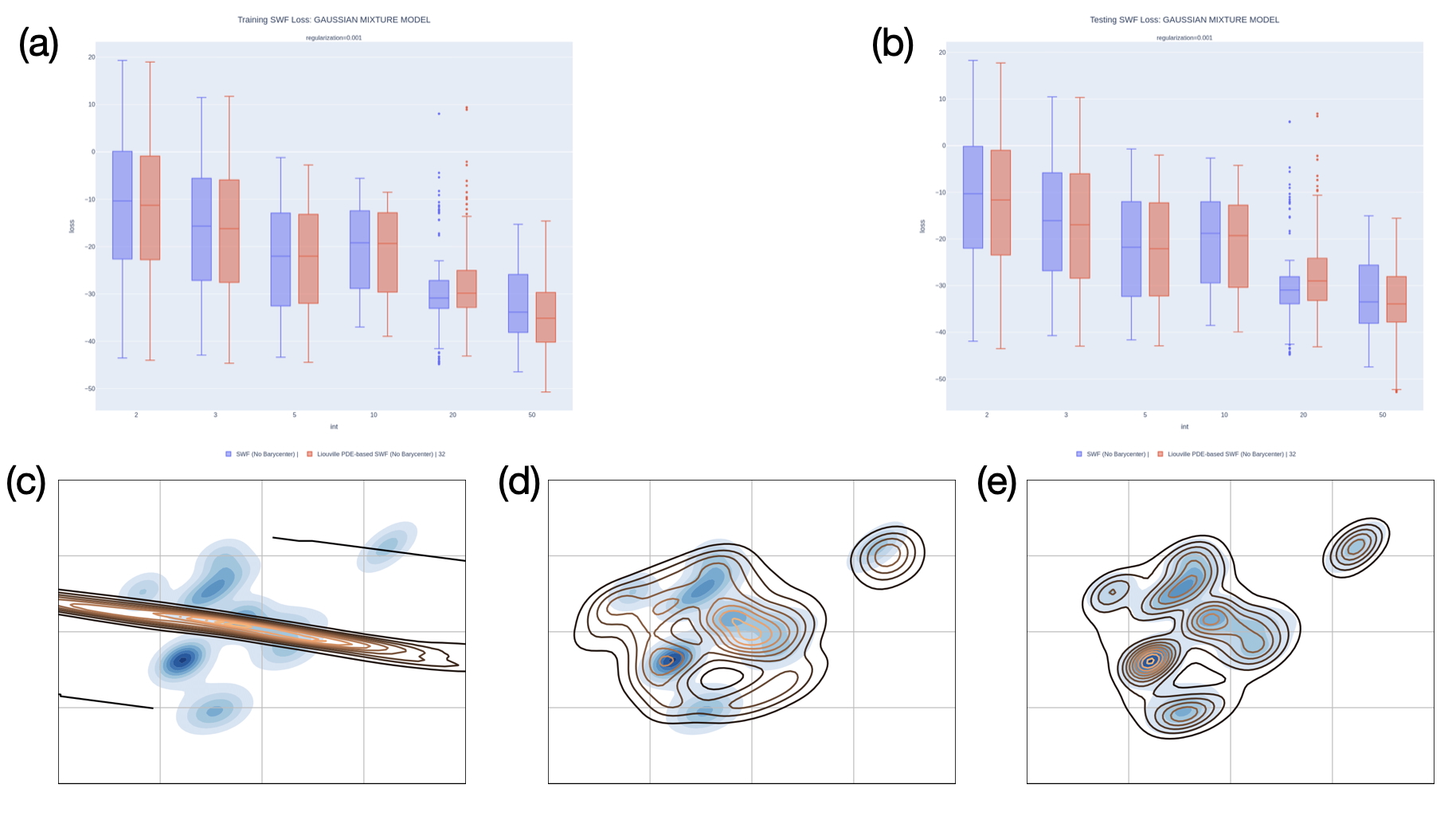}
                     \caption{\textbf{Sliced Wasserstein flows (SWF)}   (a) and (b) Training and testing the optimal transport to the defined target distribution. Two approaches are compared: the vanilla SWF  \citep{liutkus2019} and the proposed Liouville PDE-based SWF. (c)-(e) The target distribution (shaded countour plot) and the initial, intermediate, and distribution of particles (isopotential lines) in the converging Liouville PDE-based SWF.} 
\label{Fig_SWF}
\end{figure*}

We perform the first set of experiments on synthetic data where we consider a standard Gaussian mixture model (GMM) with 10 components and random parameters. Centroids are taken as sufficiently distant from each other to make the problem more challenging. We generate $P=50000$ data samples in each experiment.
In our first experiment, we set $d=2$ for visualization purposes and illustrate the general behavior of the algorithm. Figure 1 shows the evolution of the particles through the iteration. Here, we set $N_{\theta}=30$, $h=1$ and $\lambda = 10^{-4}$. We first observe that the SW cost between the empirical distributions of training data and particles is steadily decreasing along the SW flow. Furthermore, we see that the quantile functions (QFs), $F_{\theta_{\sharp}^* \bar{\mu}_{kh}^N}$ that are computed with the initial set of particles (the training stage) can be perfectly re-used for new unseen particles in a subsequent test stage, yielding similar -yet slightly higher - SW cost. The performances between Liouville PDE-based SWF and the vanilla SWF are compared in Figure \ref{Fig_SWF} and Table \ref{convergence_variance}. The loss function is Eq. (\ref{SW_min}).

\subsection{Case Study 2: Sliced Wasserstein flow for barycenter}
The setting is similar to Case Study 1, and the barycenter from two GMM distributions is computed from the vanilla SWF and the Liouville PDE-based SWF (Fig. 2). The loss function is Eq. (\ref{SW_barycenter_min}). 
\begin{figure*}[bth]
      \centering
                         \includegraphics[width=0.7\textwidth]{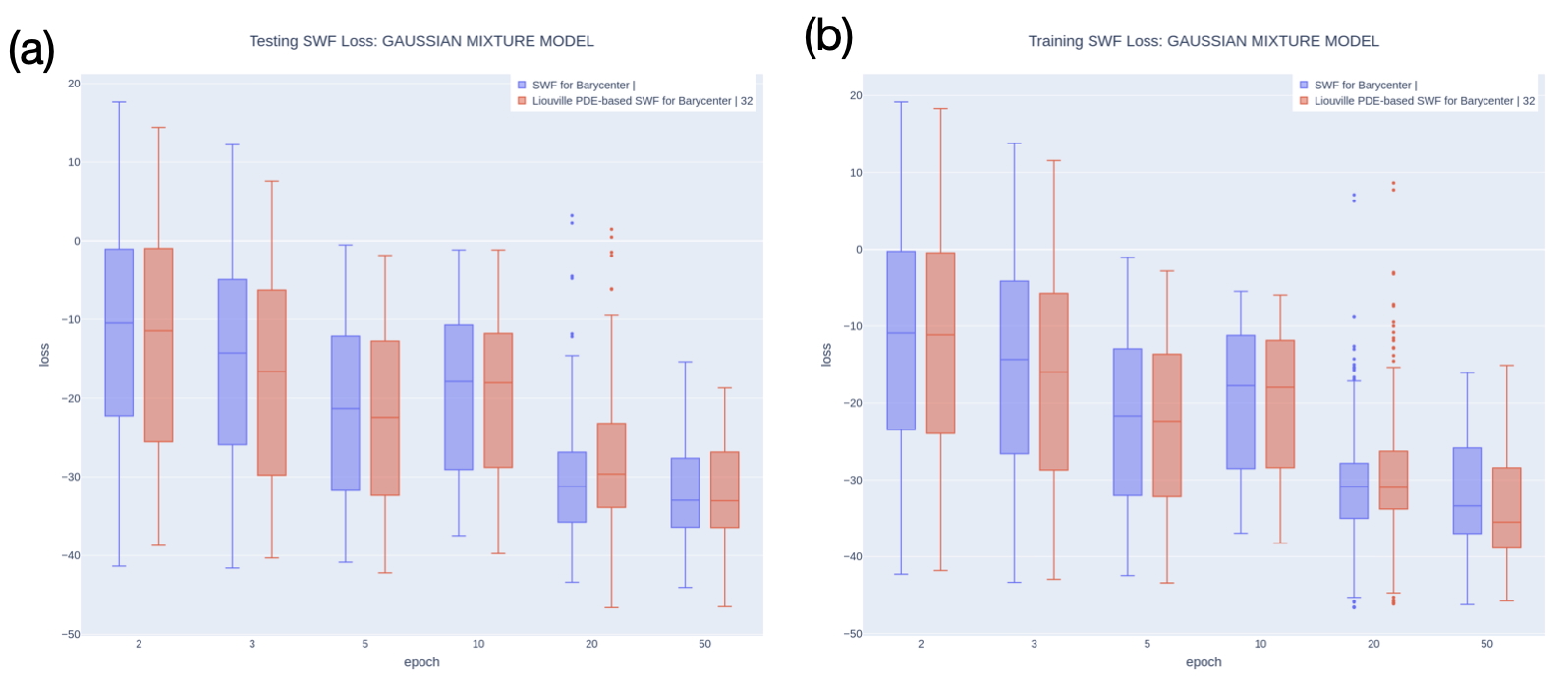}                                   
                     \caption{Sliced Wasserstein barycenter flows (SWF barycenter versus Liouville PDE-based SWF barycenter) with 2 Gaussian Mixture Models (GMM).} \label{Fig_SWF_barycenter}
    \end{figure*}

\subsection{Case Study 3: Scalability with imaging data, MNIST and CelebA}

Table \ref{image_FID} shows Frechet Inception Distance (FID), and Tables \ref{convergence_image} and \ref{image_MSE_KS} show training and testing losses at convergence and for SWF (barycenter) and Liouville PDE-based SWF (barycenter) using the best-performing configurations from the MNIST and CelebA datasets. 
Boldface indicates the lower FID and loss/variance within each training or testing comparison. Liouville PDE-based SWF (barycenter) has moderately improved SWF (barycenter) in the convergence of loss and/or variance as well as MSE and/and KS. In regards to the image generation metric (FID), Liouville PDE-based SWF has slightly outperformed SWF.
%
\begin{table}[!htbp]
\small
\centering
\begin{tabular}{|c|c|c|c|c|c|}
\hline
\textbf{Baseline (MINST)} & \textbf{Sensitive Attr.} 
& \textbf{FID} 
& \textbf{Execution time (sec)} 
& \textbf{Peak memory (MB)} \\
\hline
SWF & --  & 825.73 &176.43 & 3719.99 \\
SWF barycenter & Digit  & 828.22 & 184.48 & 3739.81 \\
Liouville SWF & --  & $\mathbf{824.98}$ & 176.04 & 3740.86 \\
Liouville SWF barycenter & Digit  & $\mathbf{819.95}$ & 183.12 & 3719.70 \\
\hline
\end{tabular}
\begin{tabular}{|c|c|c|c|c|c|}
\hline
\textbf{Baseline (CelebA)} & \textbf{Sensitive Attr.} 
& \textbf{FID} 
& \textbf{Execution time (sec)} 
& \textbf{Peak memory (MB)} \\
\hline
SWF & --  & 9076.16 & 377.43 & 4292.73 \\
SWF barycenter & Age  & 9108.81& 393.00 & 4301.20 \\
SWF barycenter & Gender  & 9073.21 & 384.53 & 4294.36 \\
Liouville SWF & --  & $\mathbf{9074.92}$ & 377.12 & 4292.60 \\
Liouville SWF barycenter & Age  & $\mathbf{9070.20}$  & 383.19 & 4297.86 \\
Liouville SWF barycenter & Gender  & $\mathbf{9073.19}$  & 395.61 & 4296.83 \\
\hline
\end{tabular}
\caption{In regards to an image generation metric, Frechet Inception Distance (FID), Liouville PDE-based SWF has slightly outperformed SWF. The execution time and peak memory are comparable to each other.}
\label{image_FID}
\end{table}
\subsection{Case Study 4: Fair regression with sliced Wasserstein barycenter}

\noindent \textbf{Communities and Crime (CRIME)} ~~~The CRIME dataset contains socio-economic, law enforcement, and crime data about communities in the US \citep{redmond2002} with 1994 examples. The task is to predict the number of violent crimes per $10^5$ population (normalized to [0,1]) with race as the protected attribute. Following \citet{calders2013}, \citealt{chzhen2020} made a binary sensitive attribute $s$ as to the percentage of the black population, which yielded 970 instances of $s=1$ with a mean crime rate 0.35 and 1024 instances of $s=-1$ with a mean crime rate 0.13. For each community (sample), the sensitive attribute PctRace is categorized among Black, White, and Asian populations with the maximum racial proportion.

\noindent \textbf{Health care spending} ~~~ \citet{Zink2019} selected a random sample of 100,000 enrollees from the IBM MarketScan Research Databases. 
Diagnosed health conditions took the form of the established Hierarchical Condition Category (HCC) variables created for risk adjustment. 
We considered the 79 HCC variables currently used in Medicare Advantage risk adjustment formulas and retained the 62 HCCs that had at least 30 enrollees with the condition. 
The sensitive attributes are age of participants (AGE), binary indicator for female (sexF), and total payments (totpay).


Liouville PDE-based SWF barycenter does mostly improve the accuracy (MSE) and/or fairness (KS) in the mean values and/or variances in comparison to the standard SWF barycenter (Table \ref{crime_health_MSE_KS}). 

\begin{figure*}[hbt]
    \centering
    \includegraphics[width=0.8 \linewidth]{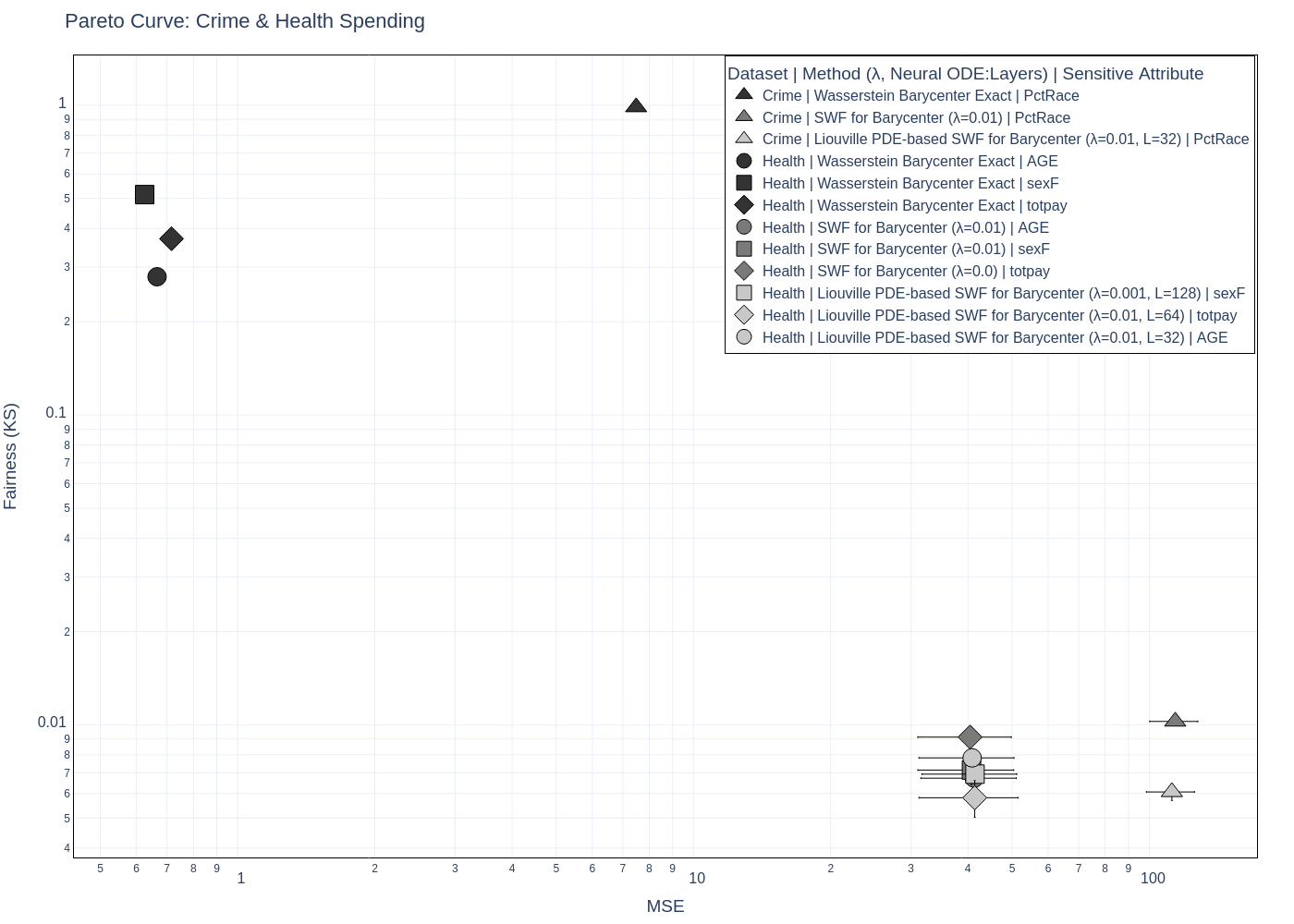}
    \caption{Pareto curves showing the accuracy (MSE)–fairness (KS) tradeoff for crime and health spending datasets. Each point corresponds to a model paired with a sensitive attribute.}
    \label{pareto_curves}
\end{figure*}

\subsection{Pareto curves of accuracy versus fairness}

The Pareto curves (Figure \ref{pareto_curves}) demonstrate exact Wasserstein barycenters attain low MSE but incur large KS disparities. Both SWF-based methods substantially reduce fairness violations, with the Liouville PDE-based SWF consistently achieving more favorable Pareto trade-offs, yielding lower KS at comparable or lower MSE across datasets and sensitive attributes. We evaluate how imposing the Liouville PDE constraint alters the fairness–accuracy tradeoff in the sliced Wasserstein Flow. 

\begin{figure}[!htbp]
\centering
    \centering
    \includegraphics[width=0.49\linewidth]{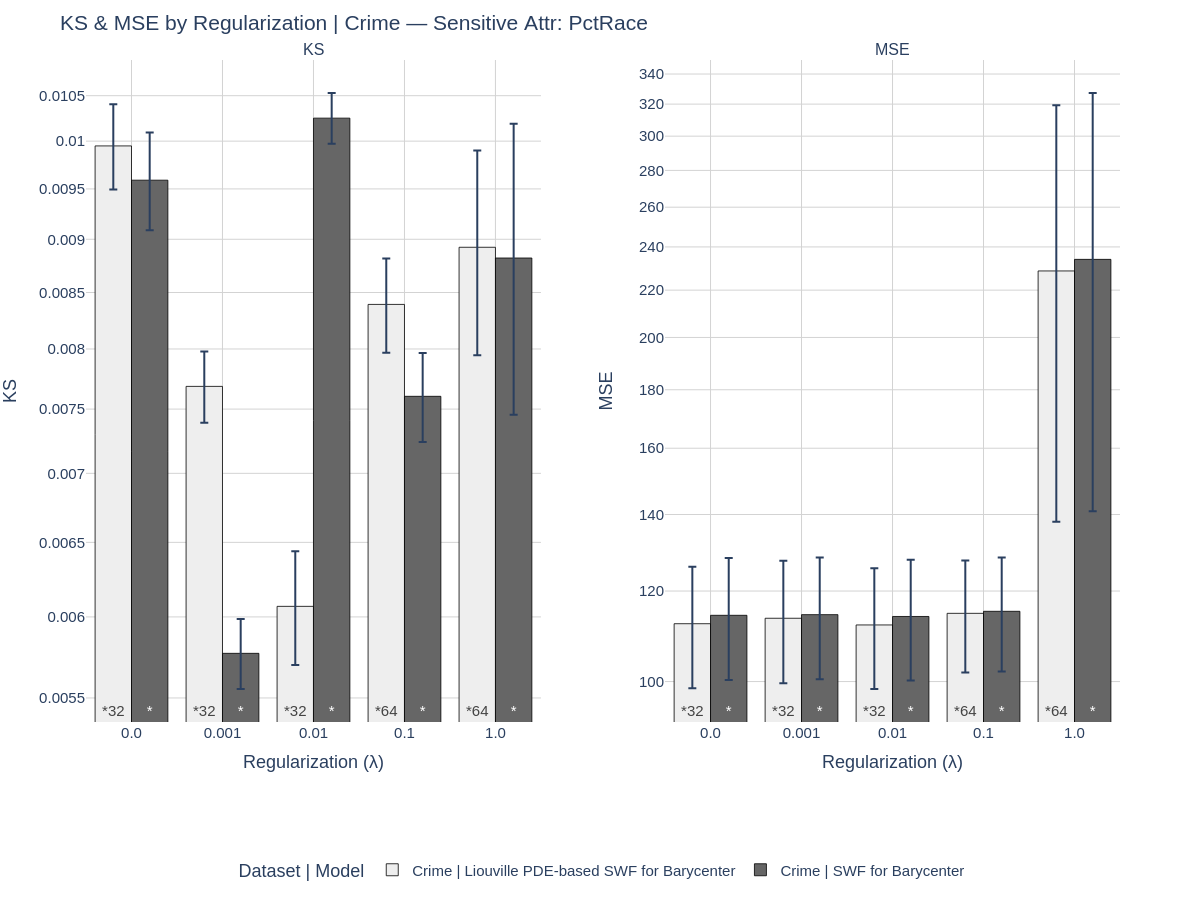}
    \includegraphics[width=0.49\linewidth]{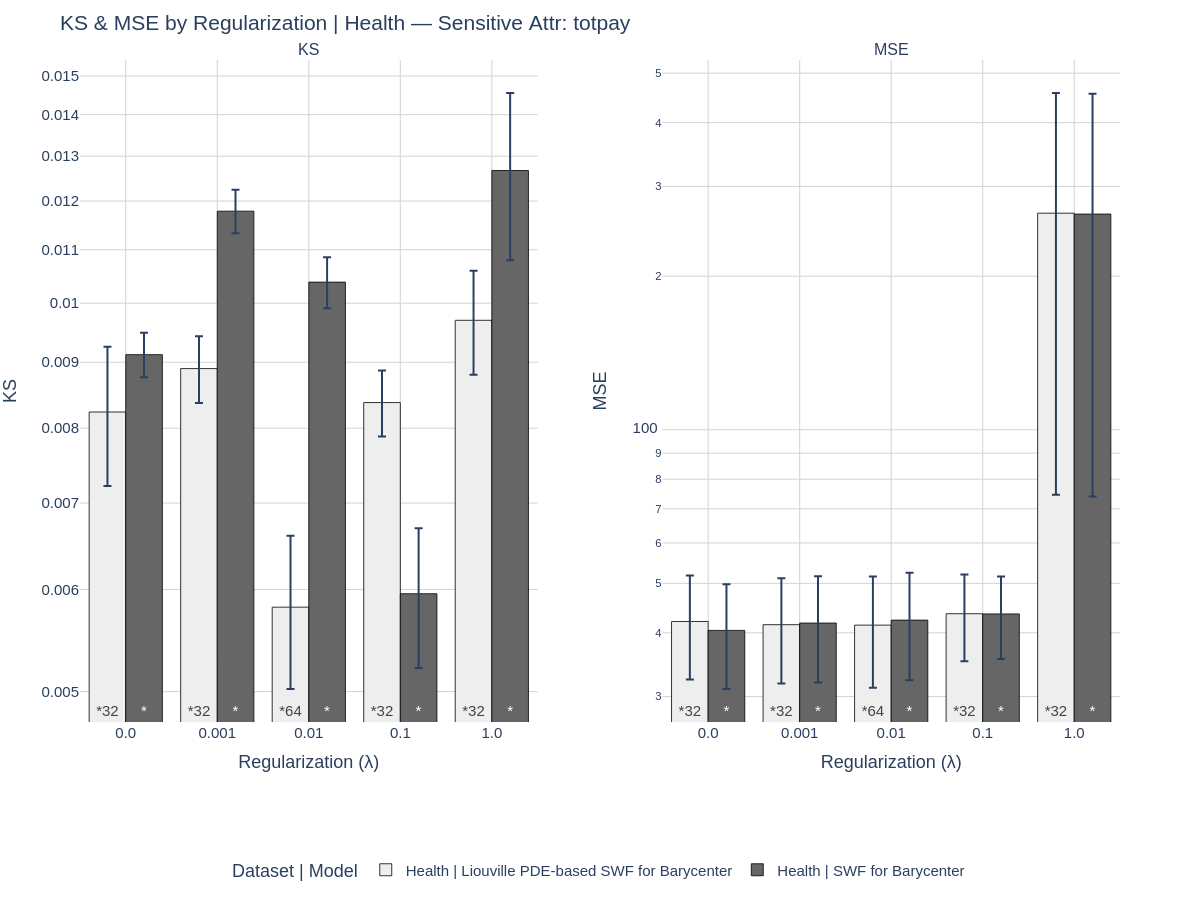}
%
\caption{
\textbf{Crime dataset (PctRace) (Left) and Health Spending dataset on \textit{totpay} (Right)} KS and MSE as functions of the regularization parameter $\lambda$ for Liouville PDE-based SWF barycenter and vanilla SWF barycenter. Error bars denote variability across runs; numbers inside bars indicate the neural ODE layer width.
\label{crime_reg_ks}
}
\end{figure}

\noindent \textbf{Communities and Crimes} ~~~ Based on Figure \ref{crime_reg_ks} (left), across regularization levels, mild regularization ($\lambda \le 0.01$) yields simultaneous reductions in KS with only marginal changes in MSE, indicating an improved accuracy--fairness trade-off relative to the unregularized case. The Liouville PDE-based SWF consistently attains lower or comparable KS than the vanilla SWF, reflecting enhanced stability on fairness under sensitive-attribute conditioning on \textit{PctRace}. As $\lambda$ increases further, both methods exhibit degraded performance, with sharp increases in MSE and no corresponding fairness gains, consistent with over-regularization. \emph{Overall, these results confirm that PDE-based regularization provides meaningful fairness improvements at low-to-moderate $\lambda$ without sacrificing predictive accuracy.}

\noindent \textbf{Health care spending cost} ~~~ Based on Figures \ref{health_care_age_reg}, \ref{health_care_sexF_reg}, and\ref{crime_reg_ks}(right), across all sensitive attributes, low-to-moderate regularization ($\lambda \le 0.1$) preserves predictive accuracy while yielding comparable or reduced KS values relative to the unregularized case, indicating a stable accuracy--fairness trade-off. The Liouville PDE-based SWF generally exhibits equal or lower KS than the vanilla SWF, particularly for \textit{AGE} and \textit{totpay}, suggesting improved control of distributional disparities under sensitive-attribute conditioning.
In contrast, strong regularization ($\lambda = 1$) leads to pronounced degradation in MSE for both methods without consistent fairness gains, reflecting over-regularization.
Overall, these results demonstrate that PDE-based regularization provides robust fairness improvements across heterogeneous health-related sensitive attributes while maintaining accuracy in the practically relevant low-$\lambda$ regime.

\begin{table}[!htbp]
\small
\centering
\begin{tabular}{|c|c|c|c|c|c|}
\hline
\textbf{Data} & \textbf{Sensitive} 
& \textbf{SWF} 
& \textbf{SWF} 
& \textbf{Liouville-SWF} 
& \textbf{Liouville-SWF} \\
 & \textbf{Attribute} 
& \textbf{training} 
& \textbf{testing} 
& \textbf{training} 
& \textbf{testing} \\
\hline
GMM & -- 
& $-26.34 \pm 11.29$
& $-26.33 \pm 11.54$
& $\textbf{-27.06} \pm 11.52$
& $\textbf{-26.70} \pm 11.55$ \\
\hline
GMM & 2 random
& $-26.47 \pm 11.38$
& $-25.95 \pm 10.75$
& $\textbf{-27.28} \pm 11.58$
& $\textbf{-27.00} \pm 11.60$ \\
\hline
Crime & -- 
& $30.91 \pm 17.12$
& $31.59 \pm \textbf{15.82}$
& \textbf{30.35 $\pm$ 16.68}
& $\textbf{31.17} \pm 16.14$ \\
\hline
Crime & PctRace 
& $31.14 \pm 16.87$
& $\textbf{31.42} \pm 15.88$
& \textbf{30.38 $\pm$ 16.84}
& $31.54 \pm \textbf{15.85}$ \\
\hline
Health & -- 
& \textbf{$23.41 \pm 16.80$}
& $23.36 \pm 18.06$
& $23.41 \pm 16.81$
& \textbf{23.29 $\pm$ 18.04} \\
\hline
Health & AGE 
& $\textbf{23.30} \pm 16.86$
& $\textbf{23.27} \pm 18.10$
& $23.41 \pm \textbf{16.81}$
& $23.29 \pm \textbf{18.04}$ \\
\hline
Health & sexF 
& $\textbf{23.26} \pm 17.06$
& $23.32 \pm \textbf{18.05}$
& $23.35 \pm \textbf{16.92}$
& \textbf{$23.32 \pm 18.09$} \\
\hline
Health & totpay 
& \textbf{$23.35 \pm 16.92$}
& $23.34 \pm 18.09$
& \textbf{23.23 $\pm$ 16.95}
& \textbf{23.21 $\pm$ 18.07} \\
\hline
\end{tabular}
\caption{Training and testing losses at convergence for vanilla SWF and Liouville-SWF using the best-performing configurations from the GMM, Crime, and Health datasets. Results are reported as mean $\pm$ standard deviation. For GMM, results are shown without and with the synthetic sensitive attribute. For Crime and Health, results are reported both without and with actual sensitive attributes. Boldface indicates the lower loss or variance within each training or testing comparison.
\label{convergence_variance}
}
\end{table}

\section{Conclusion}
The Liouville PDE-based sliced-Wasserstein flows provide improved convergence with reduced loss and/or reduced variance as a reasonable alternative to the SWF. In the high-dimensionality of around 60 and 120, the Liouville-PDE based SWF barycenter is alternatively competent to accuracy and/or fairness in comparison to the SWF, and also significantly improves in fairness in comparison to the exact Wasserstein barycenter, where computational complexity exceedingly elevates.

\noindent When the exact Wasserstein barycenter is computationally heavy, the approximate sliced-Wasserstein barycenter from Liouville PDE is useful broadly in high-dimensional statistics. The Pareto curves of accuracy and fairness in high-dimension with sensitive attributes from Liouville PDE will provide diverse aspects of decision making with rationalized fairness. In characterizing the Pareto curves in high dimensional distributions, two showcases of community crimes and health care spending costs are considered. The competence of Liouville PDE-based improvement in convergence and reduction in variance needs more exploration from diverse datasets. Moreover, the theoretical analysis of stability and convergence (e.g., exponential convergence rate) of the Liouville PDE-based SWF remains for the future. The proposed framework is expected to be extensible to general domains. One future work is the extension of Liouville PDE-based SWF to a probability space on covariance matrices, i.e., a class of Riemannian manifolds. Empirically and theoretically, the convergence of the Liouville PDE-based SWF is intriguing, where the involved neural ODE is not in Euclidean space.

\section*{Acknowledgements}
JC and PL were partially supported by NIH Data Science and NIMHD grant, 3U54MD013376-04S3. PL is partially supported by Center for Equitable AI and Machine Learning Systems (CEAMLS) Affilated Project (ID: 10072401).

\bibliography{Liouville_PDE_SWF}
\bibliographystyle{neurips_2026}


%
%
%
%
%
%
%
%
%
%

\newpage

\appendix

\section{Appendix: Proof of Theorem 2}

\setcounter{page}{1}
\setcounter{equation}{0}
\setcounter{theorem}{0}
\setcounter{definition}{0}
\setcounter{algorithm}{0}
\setcounter{figure}{0}
\setcounter{table}{0}

\renewcommand{\theequation}{S\arabic{equation}}
\renewcommand{\thealgorithm}{S\arabic{algorithm}}
\renewcommand{\thetheorem}{S\arabic{theorem}}
\renewcommand{\thedefinition}{S\arabic{definition}}
\renewcommand{\thetable}{S\arabic{table}}
\renewcommand{\thefigure}{S\arabic{figure}}

The proof is an extension for sliced-Wasserstein barycenter based on Liutkus, et al. \cite{liutkus2019}. We first need to generalize Bonnotte, et al. (Bonnotte, 2013) [Lemma 5.4.3] to distribution $\rho \in L^{\infty}(\bar{B}(0,r)), r > 0$.

\begin{theorem}
\label{thm:bigtheorem}
Let $\nu$ be a probability measure on $\bar{B}(0,1)$ with a strictly positive smooth density. Fix a time step $h > 0$, regularization constant $\lambda > 0$ and a radius $r > \sqrt{d}$. For any probability measure $\mu_0$ on $\bar{B}(0,r)$ with density $\rho_0 \in L^{\infty}(\bar{B}(0,r))$, there is a probability measure $\mu$ on $\bar{B}(0,r)$ minimizing:
\begin{eqnarray}
\mathcal{G}(\mu) = \mathcal{F}_{\lambda}^{\{\nu_s\}}(\mu) + \frac{1}{2h} \mathcal{W}_2^2(\mu, \mu_0), \nonumber
\end{eqnarray}
where $\mathcal{F}_{\lambda}^{\{\nu_s\}}$ is given by (\ref{SW_barycenter_min}). Moreover the optimal $\mu$ has a density $\rho$ on $\bar{B}(0,r)$ and:
\begin{eqnarray}
| \rho |_{L^{\infty}} \leq (1 + h / \sqrt{d})^d | \rho_0 |_{L^{\infty}}.
\end{eqnarray}
\end{theorem}
\begin{proof} 
The set of measures supported on $\bar{B}(0,r)$ is compact in the topology given by $\mathcal{W}_2$ metric. Furthermore by Ambrosio et al \cite{Ambrosio2008} [Lemma 9.4.3] $\mathcal{H}$ is lower semicontinuous on $(\mathcal{P}(\bar{B}(0,r)), \mathcal{W}_2)$. Since by (Bonnotte, 2013)[Proposition 5.1.2, Proposition 5.1.3], $\mathcal{SW}_2$ is a distance on $\mathcal{P}(\bar{B}(0,r))$, dominated by $d^{-1/2} \mathcal{W}_2$, we have:
\begin{eqnarray}
\vert \mathcal{SW}_2(\pi_0, \nu) - \mathcal{SW}_2(\pi_1, \nu) \vert \leq \mathcal{SW}_2(\pi_0, \pi_1) \leq \frac{1}{\sqrt{d}} \mathcal{W}_2(\pi_0, \pi_1). \nonumber
\end{eqnarray}
The above means that $\mathcal{SW}_2(\cdot, \nu)$ is continuous with respect to topology given by $\mathcal{W}_2$, which implies that $\mathcal{SW}_2^2(\cdot, \nu)$ is continuous in this topology as well. Therefore $\mathcal{G}: \mathcal{P}(\bar{B}(0,r)) \rightarrow (-\infty, + \infty]$ is a lower semicontinuous function on the compact set $(\mathcal{P}(\bar{B}(0,r)), \mathcal{W}_2)$. Hence there exists a mininum $\nu$ of $\mathcal{G}$ on $\mathcal{P}\bar{B}(0,r))$. Furthermore, since $\mathcal{H}(\pi) = + \infty$ for measures $\pi$ that do not admit a density with respect to Lebesgue measure, the measure $\mu$ must admit a density $\rho$.\\
If $\rho_0$ is smooth and positive on $\bar{B}(0,r)$, the inequality $S1$ is true by (Bonnotte, 2013)[Lemma 5.4.3.] When $\rho_0$ is just in $L^{\infty}(\bar{B}(0,r))$, we proceed by smoothing. For $t \in (0, 1]$, let $\rho_t$ be a function obtained by convolution of $\rho_0$ with a Gaussian kernel $(t,x,y) \mapsto (2 \pi)^{d/2} \exp(| x - y |^2 /2)$, restricting the result to $\bar{B}(0,r)$ and normalizing to obtain a probability density. Then $(\rho_t)_t$ are smooth positive densities, and it is easy to see that $\lim_{t \rightarrow 0} | \rho_0 |_{L^{\infty}} \leq | \rho_t |_{L^{\infty}}$. Furthermore, if we denote by $\mu_t$ the measure on $\bar{B}(0,r)$ with density $\rho_t$, then $\mu_t$ converge weakly to $\mu_0$. For $t \in (0,1]$ let $\hat{\mu}_t$ be the minimum of $\mathcal{F}_{\lambda}^{\{\nu_s\}}(\cdot) + \frac{1}{2h} \mathcal{W}_2^2(\cdot, \mu_t)$, and let $\hat{\rho}_t$ be the density of $\hat{\mu}_t$. Using Bonnotte, et al. (Bonnotte, 2013)[Lemma 5.4.3.] we get
\begin{eqnarray}
| \hat{\rho}_t | _{L^{\infty}} \leq (1 + h \sqrt{d})^d | \rho_t |_{L^{\infty}}, \nonumber
\end{eqnarray}
so $\hat{\rho}_t$ lies in a ball of finite radius in $L^{\infty}$. Using compactness of $\mathcal{P}(\bar{B}(0,r))$ in weak topology and compactness of closed ball in $L^{\infty}(\bar{B}(0,r))$ in weak star topology, we can choose a subsequence $\hat{\mu}_{t_k}, \hat{\rho}_{t_k}, \lim_{k \rightarrow + \infty} t_k = 0$, that converges along that subsequence to limits $\hat{\mu}$, $\hat{\rho}$. Obviously $\hat{\rho}$ is the density of $\hat{\mu}$, since for any continuous function $f$ on $\bar{B}(0,r)$ we have:
\begin{eqnarray}
\int \hat{\rho} f dx = \lim_{k \rightarrow \infty} \int \rho_{t_k} fdx = \lim_{k \rightarrow \infty} \int f d \mu_{t_k} = \int f d \mu. \nonumber
\end{eqnarray}
Furthermore, since $\hat{\rho}$ is the weak star limit of a bounded subsequence, we have:
\begin{eqnarray}
| \hat{\rho} |_{L^{\infty}} \leq \limsup_{k \rightarrow \infty} (1+h\sqrt{d})^d | \rho_{t_k} |_{L^{\infty}} \leq (1+h \sqrt{d})^d | \rho_0 |_{L^{\infty}}. \nonumber
\end{eqnarray}
To finish, we just need to prove that $\hat{\mu}$ is a minimum of $\mathcal{G}$. We remind our reader, that we already established existence of some minimum $\mu$ (that might be different from $\hat{\mu}$). Since $\hat{\mu}_{t_k}$ converges weakly to $\hat{\mu}$ in $\mathcal{P}(\bar{B}(0,r))$, it implies convergence in $\mathcal{W}_2$ as well since $\bar{B}(0,r)$ is compact. Similarly $\mu_{t_k}$ converges to $\mu_0$ in $\mathcal{W}_2$. Using the lower semicontinuity of $\mathcal{G}$ we now have:
\begin{eqnarray}
\mathcal{F}_{\lambda}^{\{\nu_s\}}(\hat{\mu}) + \frac{1}{2h} \mathcal{W}_2^2(\hat{\mu}, \mu_0) &\leq& \liminf_{k \rightarrow \infty} ( \mathcal{F}_{\lambda}^{\{\nu_s\}}(\hat{\mu}_{t_k}) + \frac{1}{2h} \mathcal{W}_2^2(\hat{\mu}_{t_k}, \mu_0)) \nonumber \\
&\leq& \liminf_{k \rightarrow \infty} \mathcal{F}_{\lambda}^{\{ \nu_s \}}(\mu) + \frac{1}{2h} \mathcal{W}_2^2(\mu, \mu_{t_k}) + \frac{1}{2h} \mathcal{W}_2^2(\hat{\mu}_{t_k}, \mu_0) - \frac{1}{2h} \mathcal{W}_2^2(\hat{\mu}_{t_k}, \mu_{t_k}) \nonumber \\
&=& \mathcal{F}_{\lambda}^{\{\nu_s\}}(\mu) + \frac{1}{2h}\mathcal{W}_2^2(\mu, \mu_0), \nonumber
\end{eqnarray}
where the second inequality comes from the fact, that $\hat{\mu}_{t_k}$ minimizes $\mathcal{F}_{\lambda}^{\{\nu_s\}}+\frac{1}{2h} \mathcal{W}_2^2(\cdot, \mu_{t_k})$. From the above inequality and previously established facts, it follows that $\hat{\mu}$ is a minimum of $\mathcal{G}$ with density satisfying S1.
\end{proof}
\begin{theorem}
Let $\nu$ be a probability measure on $\bar{B}(0,1)$ with a strictly positive smooth density. Fix a regularization constant $\lambda > 0$ and radius $r > \sqrt{d}$. Given an absolutely continuous measure $\mu_0 \in \mathcal{P}(\bar{B}(0,r))$ with density $\rho_0 \in L^{\infty}(\bar{B}(0,r))$, there is a generalized minimizing movement scheme $(\mu_t)_t$ in $\mathcal{P}(\bar{B}(0,r))$ starting from $\mu_0$ for the functional defined by
\begin{eqnarray}
\mathcal{F}^{\{\nu_s\}}(h, \mu_+, \mu_-) = \mathcal{F}_{\lambda}^{\{\nu_s\}}(\mu_+) + \frac{1}{2h} \mathcal{W}^2_2(\mu_+, \mu_-).
\end{eqnarray}
Moreover for any time $t > 0$, the probability measure $\mu_t = \mu(t)$ has density $\rho_t$ with respect to the Lebesgue measure and:
\begin{eqnarray}
| \rho_t |_{L^{\infty}} \leq e^{dt \sqrt{d}} | \rho_0 |_{L^{\infty}}.
\end{eqnarray}
\end{theorem}
\begin{proof}
We start by noting, that by S1 for any $h > 0$ there exists a piecewise constant trajectory $\mu^h$ for S2 starting at $\mu_0$. Furthermore for $t \geq 0$ measure $\mu_t^h = \mu^h(t)$ has density $\rho_t^h$, and:
\begin{eqnarray}
| \rho_t^h |_{L^{\infty}} \leq e^{d \sqrt{d} (t+h)} | \rho_0 |_{L^{\infty}}.
\end{eqnarray}
Let us choose $T>0$. We denote $\rho^h(t,x) = \rho_t^h(x)$. For $h \leq 1$, the functions $\rho^h$ lie in a ball in $L^{\infty}([0,T] \times \bar{B}(0,r))$, so from Banach-Alaoglu theorem there is a sequence $h_n$ converging to 0, such that $\rho^{h_n}$ converges in weak-star topology in $L^{\infty}([0,T] \times \bar{B}(0,r))$ to a certain limit $\rho$. Since $\rho$ has to be nonnegative except for a set of measure zero, we assume $\rho$ is nonnegative. We denote $\rho_t(x) = \rho(t,x)$. We will prove that for almost all $t, \rho_t$ is a probability density and $\mu_t^{h_n}$ converges in $\mathcal{W}_2$ to a measure $\mu_t$ with density $\rho_t$.

First of all, for almost all $t \in [0,T]$, $\rho_t$ is a probability density, since for any Borel set $A \subseteq [0,T]$ the indicator of set $A \times \bar{B}(0,r)$ is integrable, and hence by definition of the weak-start topology:
\begin{eqnarray}
\int_{A}\int_{\bar{B}(0,r)} \rho_t(x)dxdt = \lim_{n \rightarrow \infty} \int_{A} \int_{\bar{B}(0,r)} \rho_t^{h_n}(x)dxdt, \nonumber
\end{eqnarray}
and so we have to have $\int \rho_t(x)dx = 1$ for almost all $t \in [0,T]$. Nonnegativity of $\rho_t$ follows from nonnegativity of $\rho$.\\
We will now prove, that for almost all $t \in [0,T]$ the measures $\mu_t^{h_n}$ converge to a measure with density $\rho_t$. Let $t \in (0,T)$, take $\delta < \min(T-t, t)$ and $\zeta \in C^1(\bar{B}(0,r))$. We have:
\begin{eqnarray}
\bigg| \int_{\bar{B}(0,r)} \zeta d\mu_t^{h_n} &-& \int_{\bar{B}(0,r)} \zeta d\mu_t^{h_m} \bigg| \leq\\ \nonumber
&&\bigg| \int_{\bar{B}(0,r)} \zeta d\mu_{t}^{h_n} - \frac{1}{2 \delta} \int_{t-\delta}^{t+\delta} \int_{\bar{B}(0,r)} \zeta d\mu_{s}^{h_n} ds \bigg| + 
\bigg| \int_{\bar{B}(0,r)} \zeta d\mu_t^{h_m} - \frac{1}{2 \delta} \int_{t-\delta}^{t+\delta} \int_{\bar{B}(0,r)} \zeta \mu_s^{h_m} ds \bigg| + \\ \nonumber
&& \bigg| \frac{1}{2 \delta} \int_{t-\delta}^{t+\delta} \int_{\bar{B}(0,r)} \zeta d\mu_s^{h_m} ds - \frac{1}{2 \delta} \int_{t-\delta}^{t+\delta} \int_{\bar{B}(0,r)} \zeta d\mu_s^{h_n} ds \bigg|.
\end{eqnarray}
Because $\mu_t^{h_n}$ have densities $\rho_t^{h_n}$ and both $\rho^{h_n}$, $\rho^{h_m}$ converge to $\rho$ in weak-star topology, the last element of the sum on the right hand side converges to zero, as $n,m \rightarrow \infty$. Next, we get a bound on the other two terms.\\

First, if we denote by $\gamma$ the optimal transport plan between $\mu_t^{h_n}$ and $\mu_s^{h_n}$, we have:
\begin{eqnarray}
\bigg| \int_{\bar{B}(0,r)} \zeta d\mu_t^{h_n} - \int_{\bar{B}(0,r)} \zeta d \mu_s^{h_n} \bigg|^2 \leq \int_{\bar{B}(0,r) \times \bar{B}(0,r)} \vert \zeta(x) - \zeta(y) \vert^2 d\gamma(x,y) 
\leq | \nabla \zeta |_{\infty}^2 \mathcal{W}_2^2(\mu_t^{h_n}, \mu_s^{h_n}).
\end{eqnarray}
In addition, for $n_t = \lfloor t / h_n \rfloor$ and $n_s = \lfloor s / h_n \rfloor$ we have $\mu_t^{h_n} = \mu_{n_t h_n}^{h_n}$ and $\mu_s^{h_n}$. For all $k \geq 0$ we have:
\begin{eqnarray}
\mathcal{W}_2^2(\mu_{kh_n}^{h_n}, \mu_{(k+1)h_n}^{h_n}) \leq 2 h_n (\mathcal{F}_{\lambda}^{\{\nu_s\}}(\mu_{kh_n}^{h_n}) - \mathcal{F}_{\lambda}^{\{\nu_s\}} (\mu_{(k+1)h_n}^{h_n})).
\end{eqnarray}
Using this result and (S6) and assuming without loss of generality $n_t \leq n_s$, from the Cauchy-Schwartz inequality we get:
\begin{eqnarray}
\mathcal{W}_2^2(\mu_t^{h_n}, \mu_s^{h_n}) &\leq& (\sum_{k=n_t}^{n_s-1} \mathcal{W}_2(\mu_{kh_n}^{h_n}, \mu_{(k+1)h_n}^{h_n}))^2 \\
&\leq& \vert n_t - n_s \vert \sum_{k=n_t}^{n_s-1} \mathcal{W}_2^2(\mu_{kh_n}^{h_n}, \mu_{(k+1)h_n}^{h_n}) \nonumber \\
&\leq& 2 h_n \vert n_t - n_s \vert (\mathcal{F}_{\lambda}^{\{\nu_s\}}(\mu_{n_t h_n}^{h_n}) - \mathcal{F}_{\lambda}^{\{\nu_s\}}(\mu_{n_s h_n}^{h_n} )) \leq 2 C (\vert t-s \vert + h_n), \nonumber
\end{eqnarray}
where we used for the last inequality, denoting $C = \mathcal{F}_{\lambda}^{\{\nu_s\}}(\mu_0) - \min_{\mathcal{P}(\bar{B}(0,r))} \mathcal{F}_{\lambda}^{\{\nu_s\}}$, that $( \mathcal{F}_{\lambda}^{\{\nu_s\}}(\mu_{k h_n}^{h_n}))_n$ is non-increasing by (S7) and $\min_{\mathcal{P}(\bar{B}(0,r))}\mathcal{F}_{\lambda}^{\{\nu_s\}}$ is finite since $\mathcal{F}_{\lambda}^{\{\nu_s\}}$ is lower semi-continuous. Finally, using Jensen's inequality, the above bound and (S6) we get:
\begin{eqnarray}
\bigg| \int_{\bar{B}(0,r)} \zeta d \mu_t^{h_n} - \frac{1}{2 \delta} \int_{t - \delta}^{t + \delta} \int_{\bar{B}(0,r)} \zeta d \mu_s^{h_n} ds \bigg|^2 \nonumber
&\leq& \frac{1}{2 \delta} \int_{t-\delta}^{t+\delta} \bigg| \int_{\bar{B}(0,r)} \zeta d \mu_t^{h_n} - \int_{\bar{B}(0,r)} \zeta d \mu_s^{h_n} \bigg|^2 ds\\ \nonumber
&\leq& \frac{C | \nabla \zeta |_{\infty}^2}{\delta} \int_{t-\delta}^{t+\delta} (| t-s | + h_n) ds\\ \nonumber
&\leq& 2 C | \nabla \zeta |_{\infty}^2 (h_n + \delta). \nonumber
\end{eqnarray}
Together with (S5), when taking $\delta = h_n$, this result means that $\int_{\bar{B}(0,r)} \zeta d \mu_t^{h_n}$ is a Cauchy sequence for all $t \in (0,T)$. On the other hand, since $\rho^{h_n}$ converges to $\rho$ in weak-star topology on $L^{\infty}$, the limit of $\int_{\bar{B}(0,r)} \zeta d \mu_t^{h_n}$ has to be $\int_{\bar{B}(0,r)} \zeta(x) \rho_t(x)dx$ for almost all $t \in (0,T)$. This means that for almost all $t \in [0,T]$ sequence $\mu_t^{h_n}$ converges to a measure $\mu_t$ with density $\rho_t$. Let $S \in [0,T]$ be the set of times such that for $t \in S$ sequence $\mu_t^{h_n}$ converges to a measure $\mu_t$. As we established almost all points from $[0,T]$ belong to $S$. Let $t \in [0,T] \backslash S$. Then, there exists a sequence of times $t_k \in S$ converging to $t$, such that $\mu_{t_k}$ converge to some limit $\mu_t$. We have:
\begin{eqnarray}
\mathcal{W}_2(\mu_t^{h_n}, \mu_t) \leq \mathcal{W}_2(\mu_t^{h_n}, \mu_{t_k}^{h_n}) + \mathcal{W}_2(\mu_{t_k}^{h_n}, \mu_{t_k}) + \mathcal{W}_2(\mu_{t_k}, \mu_t). \nonumber
\end{eqnarray}
From which we have for all $k \geq 1$:
\begin{eqnarray}
\limsup_{n \rightarrow \infty} \mathcal{W}_2( \mu_t^{h_n}, \mu_t) \leq \mathcal{W}_2(\mu_{t_k}, \mu_{t}) + \limsup_{n \rightarrow \infty} \mathcal{W}_2(\mu_t^{h_n}, \mu_{t_k}^{h_n}), \nonumber
\end{eqnarray}
and using (S8), we get $\mu_t^{h_n} \rightarrow \mu_t$. Furthermore, the measure $\mu_t$ has to have density, since $\rho_t^{h_n}$ lies in a ball in $L^{\infty}(\bar{B}(0,r))$, so we can choose a subsequence of $\rho_t^{h_n}$ converging in weak-star topology to a certain limit $\hat{\rho}_t$, which is the density of $\mu_t$. We use now the diagonal argument to get convergence for all $t > 0$. Let $(T_k)_{k=1}^{\infty}$ be a sequence of times increasing to infinity. Let $h_n^1$ be a sequence converging to 0, such that $\mu_t^{h_n^1}$ converge to $\mu_t$ for all $t \in [0,T_1]$. Using the same arguments as above, we can choose a subsequence $h_n^2$ of $h_n^1$, such that $\mu_t^{h_n^2}$ converges to a limit $\mu_t$ for all $t \in [0,T_2]$. Inductively, we construct subsequences $h_n^k$, and in the end take $h_n=h_n^n$. For this subsequence we have that $\mu_t^{h_n}$ converges to $\mu_t$ for all $t > 0$, and $\mu_t$ has a density satisfying the bound from the statement of the theorem.\\
Finally, note that (S2) follows from (S4).
\end{proof}

\begin{theorem}
Let $(\mu_t)_{t \geq 0}$ be a generalized minimizing movement scheme given by Theorem S3 with initial distribution $\mu_0$ with density $\rho_0 \in L(\bar{B}(0,r))$. We denote by $\rho_t$ the density of $\mu_t$ for all $t \geq 0$. Then $\rho_t$ satisfies the continuity equation:
\begin{eqnarray}
\frac{\partial \rho_t}{\partial t} + \nabla \cdot (v_t \rho_t) + \lambda \Delta \rho_t = 0,~~~~~ v_t(x) = -\int_{\mathbb{S}^{d-1}} \sum_{s} p_s \psi'_{s, t, \theta} ( \langle x,\theta \rangle) \theta d \theta, \nonumber
\end{eqnarray}
 in a weak sense, that is for all $\xi \in C_c^{\infty}([0, \infty) \times \bar{B}(0,r))$ we have:
\begin{eqnarray}
\int_0^{\infty} \int_{\bar{B}(0,r)} [ \frac{\partial \xi}{\partial t}(t,x) - v_t \nabla \xi(t,x) - \lambda \Delta \xi(t,x) ] \rho_t(x) dxdt = - \int_{\bar{B}(0,r)} \xi(0,x) \rho_0(x)dx. \nonumber
\end{eqnarray}
\end{theorem}
\begin{proof}
Our proof is based on the proof of \cite{bonnotte2013} [Theorem 5.6.1]. We proceed in five steps.\\
(1) Let $h_n \rightarrow 0$ be a sequence given by Theorem S3, such that $\mu_t^{h_n}$ converges to $\mu_t$ pointwise. Furthermore we know that $\mu^{h_n}$ have densities $\rho^{h_n}$ that converge to $\rho$ in $L^r$, for $r \geq 1$, and in weak-star topology in $L^{\infty}$. Let $\xi \in C_c^{\infty}([0, \infty) \times \bar{B}(0,r))$. We denote $\xi_k^n(x) = \xi(k h_n, x)$. Using Part 1 of the proof of (Bonnotte, 2013) [Theorem 5.6.1], we obtain:
\begin{eqnarray}
\int_{\bar{B}(0,r)} \xi(0,x) \rho_0(x) dx &+& \int_0^{\infty} \int_{\bar{B}(0,r)} \frac{\partial \xi}{\partial t} (t,x) \rho_t(x) dxdt \nonumber \\
&=& \lim_{n \rightarrow \infty} -h_n \sum_{k=1}^{\infty} \int_{\bar{B}(0,r)} \xi_k^n(x) \frac{\rho_{kh_n}^{h_n}(x) - \rho_{(k-1)h_n}^{h_n}(x)}{h_n}dx. 
\end{eqnarray}
(2) Again, this part is the same as Part 2 of the proof of (Bonnotte, 2013)[Theorem 5.6.1]. For any $\theta \in \mathbb{S}^{d-1}$ we denote by $\psi_{t, \theta}$ the unique Kantorovich potential from $\theta_{\sharp}^* \mu_t$ to $\theta_{\sharp}^* \nu$, and by $\psi_{t, \theta}^{h_n}$ the unique Kantorovich potential from $\theta_{\sharp}^* \mu_t^{h_n}$ to $\theta_{\sharp}^* \nu$. Then, by the same reasoning as Part 2 of the proof of (Bonnotte, 2013) [Theorem 5.6.1], we get:
\begin{eqnarray}
&&\int_0^{\infty} \int_{\bar{B}(0,r)} \int_{\mathbb{S}^{d-1}} \sum_s p_s (\psi_{s, t, \theta}' ( \langle \theta, x \rangle)  \langle \theta, \nabla \xi(x,t) \rangle d \theta d \mu_t(x)dt \nonumber \\
&=& \lim_{n \rightarrow \infty} h_n \sum_{k=1}^{\infty} \int_{\bar{B}(0,r)} \int_{\mathbb{S}^{d-1}} \sum_s p_s \psi_{s, kh_n, \theta}^{h_n} (\theta^*) \langle \theta, \nabla \xi_k^n \rangle d\theta d\mu_{k h_n}^{h_n}.
\end{eqnarray}
(3) Since $\xi$ is compactly supported and smooth, $\Delta \xi$ is Lipschitz, and so for any $t \geq 0$ if we take $k=\lfloor t/h_n \rfloor$ we get $\vert \Delta \xi_k^n(x) - \Delta \xi(t,x) \vert \leq C h_n$ for some constant $C$. Let $T>0$ be such that $\xi(t,x)=0$ for $t > T$. We have:
\begin{eqnarray}
\bigg| \sum_{k=1}^{\infty} h_n \int_{\bar{B}(0,r)} \Delta \xi_k^n(x) \rho_{k h_n}^{h_n} (x) dx - \int_0^{+ \infty} \int_{\bar{B}(0,r)} \Delta \xi(t,x) \rho_t^{h_n} (x) dxdt \bigg| \leq CTh_n. \nonumber
\end{eqnarray}
On the other hand, we know, that $\rho^{h_n}$ converges to $\rho$ in weak star topology on $L^{\infty}([0,T] \times \bar{B}(0,r))$, and $\Delta \xi$ is bounded, so:
\begin{eqnarray}
\lim_{n \rightarrow +\infty} \bigg| \int_0^{+ \infty} \int_{\bar{B}(0,r)} \Delta \xi(t,x) \rho_t^{h_n} (x) dxdt - \int_0^{+ \infty} \int_{\bar{B}(0,r)} \Delta \xi(t,x) \rho_t(x) dxdt \bigg| = 0. \nonumber
\end{eqnarray}
Combining those two results give:
\begin{eqnarray}
\lim_{n \rightarrow \infty} h_n \sum_{k=1}^{\infty} \int_{\bar{B}(0,r)} \Delta \xi_k^n(x) \rho_{k h_n}^{h_n} (x) dx = \int_0^{+ \infty} \int_{\bar{B}(0,r)} \Delta \xi(t,x) \rho_t(x)dxdt.
\end{eqnarray}
(4) Let $\phi_k^{h_n}$ denote the unique Kantorovich potential from $\mu_{kh_n}^{h_n}$ to $\mu_{(k-1)h_n}^{h_n}$. Using Bonnote, et al. \cite{bonnotte2013} [Propositions 1.5.7 and 5.1.7], as well as Jordan, et al. \cite{jordan1998} [Equation (38)] with $\Psi = 0$, and optimality of $\mu_{k h_n}^{h_n}$, we get:
\begin{eqnarray}
\frac{1}{h_n} \int_{\bar{B}(0,r)} \langle \nabla \phi_k^{h_n}(x), \nabla \xi_k^n(x) \rangle d \mu_{k h_n}^{h_n} (x) &-& \int_{\bar{B}(0,r)} \int_{\mathbb{S}^{d-1}} \sum_s p_s (\psi_{s, k h_n}^{h_n})' (\theta^*) \langle \theta, \nabla \xi_k^n(x) \rangle d \theta d \mu_{k h_n}^{h_n} (x) \nonumber \\
- \lambda \int_{\bar{B}(0,r)} \Delta \xi_k^n(x) d \mu_{k h_n}^{h_n} (x),
\end{eqnarray}
which is the derivative of $\mathcal{F}_{\lambda}^{\{\nu_s\}} (\cdot) + \frac{1}{2 h_n} \mathcal{W}_2^2(\cdot, \mu_{(k-1)h_n})$ in the direction given by vector field $\nabla \xi_k^n$ is zero.\\
Let $\gamma$ be the optimal transport between $\mu_{k h_n}^{h_n}$ and $\mu_{(k-1)h_n}^{h_n}$. Then:
\begin{eqnarray}
\int_{\bar{B}(0,r)} \xi_k^n(x) \frac{\rho_{kh_n}^{h_n} (x) - \rho_{(k-1)h_n}^{h_n} (x)}{h_n}dx &=& \frac{1}{h_n} \int_{\bar{B}(0,r)} ( \xi_k^n(y) - \xi_k^n(x)) d \gamma(x,y),\\
\frac{1}{h_n} \int_{\bar{B}(0,r)} \langle \nabla \phi_k^{h_n} (x), \nabla \xi_k^n(x) \rangle d \mu_{k h_n}^{h_n} (x) &=& \frac{1}{h_n} \int_{\bar{B}(0,r)} \langle \nabla \xi_k^n(x), y-x \rangle d \gamma(x,y).
\end{eqnarray}
Since $\xi$ is $C_c^{\infty}$, it has Lipschitz gradient. Let $C$ be twice the Lipschitz contant of $\nabla \xi$. Then we have $\vert \xi(y) - \xi(x) - \langle \nabla \xi(x), y-x \rangle \vert \leq C \vert x-y \vert ^2$, and hence
\begin{eqnarray}
\int_{\bar{B}(0,r)} \vert \xi_k^n(y) - \xi_k^n(x) - \langle \nabla \xi_k^n(x), y-x \rangle \vert d\gamma (x,y) \leq C \mathcal{W}_2^2 ( \mu_{(k-1)h_n}^{h_n}, \mu_{k h_n}^{h_n}).
\end{eqnarray}
Combining (S13), (S14) and (S15), we get:
\begin{eqnarray}
\bigg| \sum_{k=1}^{\infty} h_n \int_{\bar{B}(0,r)} \xi_k^n(x) \frac{\rho_{k h_n}^{h_n} - \rho_{(k-1)h_n}^{h_n}}{h_n} dx &+& \sum_{k=1}^{\infty} h_n \int_{\bar{B}(0,r)} \langle \nabla \phi_k^{h_n}, \nabla \xi_k^n \rangle d\mu_{kh_n}^{h_n} \bigg| \nonumber \\
\leq C \sum_{k=1}^{\infty} \mathcal{W}_2^2 ( \mu_{(k-1)h_n}^{h_n}, \mu_{kh_n}^{h_n}).
\end{eqnarray}
As some $\mathcal{F}_{\lambda}^{\{\nu_s\}}$ have a finite minimum on $\mathcal{P}(\bar{B}(0,r))$, we have:
\begin{eqnarray}
\sum_{k=1}^{\infty} \mathcal{W}_2^2 (\mu_{(k-1)h_n}^{h_n}, \mu_{k h_n}^{h_n}) &\leq& 2 h_n \sum_{k=1}^{\infty} \mathcal{F}_{\lambda}^{\{\nu_s\}} ( \mu_{(k-1)h_n}^{h_n}) - \mathcal{F}_{\lambda}^{\{\nu_s\}}(\mu_{k h_n}^{h_n}) \nonumber \\
&\leq& 2 h_n ( \mathcal{F}_{\lambda}^{\{\nu_s\}} (\mu_0) - \min_{\mathcal{P}(\bar{B}(0,r))} \mathcal{F}_{\lambda}^{\{\nu_s\}}),
\end{eqnarray}
and so the sum on the right hand side of the equation goes to zero as $n$ goes to infinity.\\
From (S16), (S17) and (S12) we conclude:
\begin{eqnarray}
&&\lim_{n \rightarrow \infty} -h_n \sum_{k=1}^{\infty} \xi_k^n(x) \frac{\rho_{k h_n}^{h_n} - \rho_{(k-1)h_n}^{h_n}}{h_n} dx = \\
&&\lim_{n \rightarrow \infty} \bigg( h_n \sum_{k=1}^{\infty} \int_{\bar{B}(0,r)} \int_{\mathbb{S}^{d-1}} \sum_s \psi_{s, kh_n, \theta}^{h_n} (\theta^*) \langle \theta, \nabla \xi_k^n \rangle d \theta d \mu_{k h_n}^{h_n} + h_n \sum_{k=1}^{\infty} \int _{\bar{B}(0,r)} \nabla \xi_k^n(x) \rho_{k h_n}^{h_n} (x)dx \bigg), \nonumber
\end{eqnarray}
where both limits exist, since the difference of left hand side and right hand side of the equation goes to zero, while the left hand side converges to a finite value by (S9).\\
(5) Combining (S9), (S10), (S11) and (S18) we get the result.
\end{proof}

\newpage

%
\begin{algorithm}[hbt]
\caption{Liouville PDE-based sliced-Wasserstein barycenter flows}
\begin{algorithmic}
\STATE{\textbf{Input}: $\mathcal{D} = \{ y_{i,s} \}_{1 \leq i \leq P_s}^{1 \leq s \leq N_s}$, $\mu_0$, $N$, $N_{\theta}$, $h$, $\lambda$}
\STATE{\textbf{Output}: $\{ \bar{X}_K^i \}_{i=1}^N$}
\STATE{$\bar{X}_o^i \sim \mu_0$, i.i.d.}
\STATE{$\theta_n \sim$ from $\mathbb{S}^{d-1}$}
\FOR{$\theta \in \{ \theta_n \}_{n=1}^{N_{\theta}}$}
\FOR{$s=1, ..., N_s$}
\STATE{$F_{\theta^*_{\#} {\nu_s}}^{-1} = {\rm QF} \{ \langle \theta, y_{i,s} \rangle \}_{i=1}^{P_s}$}
\ENDFOR
\ENDFOR
\FOR{$k=0, ..., K-1$}
\FOR{$\theta \in \{ \theta_n \}_{n=1}^{N_{\theta}}$} 
\STATE{$F_{\theta^*_{\#} \bar{\mu}_{kh}^N} = {\rm CDF} \{ \langle \theta, \bar{X}_k^i \rangle \}_{i=1}^N$}
\ENDFOR
\STATE{$\tilde{v}_k(\bar{X}_k^i) = -\frac{1}{N_{\theta}} \sum_{n=1}^{N_{\theta}} \sum_{s \in S} p_s \psi_{s,k,\theta_n}' ( \langle \theta_n, \bar{X}_k^i \rangle) \theta_n$} 
\STATE{$ ~~~~~~~~~~~~~~~~~~ - \lambda \nabla \log \rho(\bar{X}_k^i,t_k)$}
\STATE{$\bar{X}_{k+1}^i = \bar{X}_k^i + h \tilde{v}_k (\bar{X}_k^i)$}
\STATE{Update the density $\log \rho(\cdot,t_k)$ from Eq. (\ref{neural_ODE}).}
\ENDFOR
\end{algorithmic}
\label{algorithm1}
\end{algorithm}
%

\section{Appendix: Hyperparameters}

\begin{enumerate}

\item Gaussian Mixture Model parameters:

\begin{itemize}
\item overall number of samples, 6000
\item number of quantiles, 50
\item dimension of input, 2
\item number of samples in training data, 5000
\item step size, 1e-3
\item number of samples in testing data, 1000
\item number of thetas for sliced Wasserstein integration, 256
\item number of epochs, 100
\end{itemize}

\item Image data MNIST parameters:

\begin{itemize}
\item Regularization:  0, 0.001, 0.01, 0.1, 1 
\item Layer size when using ODE - 64, 128, or 256
\item Input dimension: 784 (28×28 flattened)
\item Training examples: 60,000
\item Sensitive attribute: column 0 only (Digit class label)
\end{itemize}

\item Image data CelebA parameters:

\begin{itemize}
\item Regularization:  0, 0.001, 0.01, 0.1, 1 
\item Layer size when using ODE - 64, 128, or 256
\item Input dimension: 3,072 (32×32×3 flattened RGB)
\item Training examples: 100,000
\item Sensitive attributes: column 21 (Gender) and column 40 (Age)
\end{itemize}

\item Crime Data parameters:

\begin{itemize}
\item overall number of samples, 1994
\item number of quantiles, 50
\item dimension of input, 122
\item number of samples in training data, 1694
\item step size, 1e-2
\item number of samples in testing data, 300
\item number of thetas for sliced Wasserstein integration, 512
\item number of epochs, 200
\end{itemize}

\item Health Data parameters:

\begin{itemize}
\item overall number of samples, 100000
\item number of quantiles, 50
\item dimension of input, 63
\item number of samples in training data, 70000
\item step size, 1e-2
\item number of samples in testing data, 30000
\item number of thetas for sliced Wasserstein integration, 256
\item number of epochs, 200
\end{itemize}

\item Adjusted depending on the data set:
\begin{itemize}
%
\item Autoencoder used within swf:
\begin{itemize}
\item criterion: Defines the loss function as Binary Cross Entropy Loss (nn.BCELoss())
\item optimizer: Uses the Adam optimizer to update the model parameters with a learning rate of 1e-3.
\end{itemize}
\end{itemize}
\end{enumerate}
Flags to set in \emph{swf$\_$liouville.py}:

\begin{enumerate}
\item barycenter: Bool to tell algorithm to use barycenter
\item fokk$\_$louiville: Bool to tell algorithm to use the Liouville PDE-based SWF
\item num$\_$points = Num$\_$samples
\item num$\_$dim = input$\_$dim
\item batch$\_$size = num$\_$test
\end{enumerate}

\section{Appendix: Supplementary results}

\subsection{Further details in the performance of Liouville PDE-based SWF and SWF barycenter}

The following table shows training and testing losses at convergence for SWF and Liouville PDE-based SWF using the best-performing configurations from the MNIST and CelebA datasets.

\begin{table}[!htbp]
\small
\caption{Results are reported as mean $\pm$ standard deviation to six significant figures. Boldface indicates the lower loss within each training or testing comparison.}
\centering
\begin{tabular}{|c|c|c|c|c|c|}
\hline
\textbf{Data} & \textbf{Sensitive} & \textbf{SWF} & \textbf{SWF} & \textbf{Liouville-SWF} & \textbf{Liouville-SWF} \\
 & \textbf{Attributes} & \textbf{training} & \textbf{testing} & \textbf{training} & \textbf{testing} \\
\hline
MNIST & -- & $47.384 \pm 18.446$ & $47.374 \pm 18.454$ & $\mathbf{47.360} \pm 18.440$ & $\mathbf{47.365} \pm 18.452$ \\
\hline
MNIST & Digit & $\mathbf{47.370} \pm 18.442$ & $\mathbf{47.360} \pm 18.484$ & $47.382 \pm 18.430$ & $47.392 \pm 18.438$ \\
\hline
CELEBA & -- & $59.340 \pm 18.408$ & $\mathbf{59.318} \pm 18.452$ & $\mathbf{59.322} \pm 18.406$ & $59.325 \pm 18.410$ \\
\hline
CELEBA & Age & $\mathbf{59.325} \pm 18.434$ & $59.316 \pm 18.446$ & $59.335 \pm 18.396$ & $\mathbf{59.315} \pm 18.443$ \\
\hline
CELEBA & Gender & $\mathbf{59.320} \pm 18.421$ & $59.331 \pm 18.435$ & $59.347 \pm 18.406$ & $\mathbf{59.318} \pm 18.451$ \\
\hline
\end{tabular}
\label{convergence_image}
\end{table}

\begin{table}[!htbp]
\small
\caption{Comparison of accuracy (MSE) and fairness (KS) between SWF barycenter and Liouville PDE-based SWF across MNIST and CELEBA datasets for each sensitive attribute.}
\centering
\begin{tabular}{|c|c|c|c|c|c|c|}
\hline
\textbf{Data} & \textbf{Baseline} & \textbf{Sensitive} & $\boldsymbol{\lambda}$ & \textbf{Neural ODE} & \textbf{MSE} & \textbf{KS} \\
    &  & Attribute &  & \textbf{Layer Size} &  &  \\
\hline
CelebA & Liouville PDE-based & AGE & 0.0 & 128 & $\mathbf{3565.32 \pm 46.40}$ & $\mathbf{0.0057 \pm 0.00019}$ \\
\hline
CelebA & SWF barycenter & AGE & 0.0 & -- & $3569.83 \pm 46.44$ & $0.0084 \pm 0.00020$ \\
\hline
CelebA & Liouville PDE-based & AGE & 0.001 & 64 & $\mathbf{3565.53 \pm 46.28}$ & $0.0078 \pm 0.000078$ \\
\hline
CelebA & SWF barycenter & AGE & 0.001 & -- & $3577.32 \pm 46.51$ & $\mathbf{0.0074 \pm 0.000057}$ \\
\hline
CelebA & Liouville PDE-based & AGE & 0.01 & 256 & $\mathbf{3571.67 \pm 46.38}$ & $\mathbf{0.0059 \pm 0.000045}$ \\
\hline
CelebA & SWF barycenter & AGE & 0.01 & -- & $3577.27 \pm 46.57$ & $0.010 \pm 0.00012$ \\
\hline
CelebA & Liouville PDE-based & AGE & 0.1 & 128 & $\mathbf{3574.66 \pm 46.43}$ & $0.011 \pm 0.00011$ \\
\hline
CelebA & SWF barycenter & AGE & 0.1 & -- & $3584.40 \pm 46.64$ & $\mathbf{0.0093 \pm 0.00010}$ \\
\hline
CelebA & Liouville PDE-based & AGE & 1.0 & 256 & $\mathbf{3562.77 \pm 45.69}$ & $\mathbf{0.0061 \pm 0.00022}$ \\
\hline
CelebA & SWF barycenter & AGE & 1.0 & -- & $3575.62 \pm 45.70$ & $0.0063 \pm 0.00019$ \\
\hline
CelebA & Liouville PDE-based & GENDER & 0.0 & 128 & $\mathbf{3572.64 \pm 46.43}$ & $0.0094 \pm 0.00017$ \\
\hline
CelebA & SWF barycenter & GENDER & 0.0 & -- & $3573.92 \pm 46.51$ & $\mathbf{0.0065 \pm 0.00015}$ \\
\hline
CelebA & Liouville PDE-based & GENDER & 0.001 & 256 & $3568.22 \pm 46.35$ & $\mathbf{0.0050 \pm 0.000050}$ \\
\hline
CelebA & SWF barycenter & GENDER & 0.001 & -- & $\mathbf{3566.99 \pm 46.39}$ & $0.0058 \pm 0.000093$ \\
\hline
CelebA & Liouville PDE-based & GENDER & 0.01 & 64 & $3575.91 \pm 46.53$ & $0.0092 \pm 0.000060$ \\
\hline
CelebA & SWF barycenter & GENDER & 0.01 & -- & $\mathbf{3570.43 \pm 46.37}$ & $\mathbf{0.0058 \pm 0.000071}$ \\
\hline
CelebA & Liouville PDE-based & GENDER & 0.1 & 128 & $3572.67 \pm 46.48$ & $\mathbf{0.0080 \pm 0.000057}$ \\
\hline
CelebA & SWF barycenter & GENDER & 0.1 & -- & $\mathbf{3571.46 \pm 46.39}$ & $0.012 \pm 0.000080$ \\
\hline
CelebA & Liouville PDE-based & GENDER & 1.0 & 256 & $3573.22 \pm 45.42$ & $\mathbf{0.0089 \pm 0.00012}$ \\
\hline
CelebA & SWF barycenter & GENDER & 1.0 & -- & $\mathbf{3567.27 \pm 45.52}$ & $0.010 \pm 0.00044$ \\
\hline
MNIST & Liouville PDE-based & DIGIT & 0.0 & 128 & $\mathbf{898.75 \pm 11.68}$ & $\mathbf{0.0087 \pm 0.00016}$ \\
\hline
MNIST & SWF barycenter & DIGIT & 0.0 & -- & $904.86 \pm 11.77$ & $0.014 \pm 0.00011$ \\
\hline
MNIST & Liouville PDE-based & DIGIT & 0.001 & 64 & $\mathbf{902.18 \pm 11.74}$ & $\mathbf{0.0059 \pm 0.000095}$ \\
\hline
MNIST & SWF barycenter & DIGIT & 0.001 & -- & $903.00 \pm 11.74$ & $0.0076 \pm 0.00011$ \\
\hline
MNIST & Liouville PDE-based & DIGIT & 0.01 & 128 & $904.37 \pm 11.75$ & $0.0085 \pm 0.000092$ \\
\hline
MNIST & SWF barycenter & DIGIT & 0.01 & -- & $\mathbf{901.56 \pm 11.73}$ & $\mathbf{0.0061 \pm 0.00028}$ \\
\hline
MNIST & Liouville PDE-based & DIGIT & 0.1 & 256 & $905.45 \pm 11.77$ & $0.0092 \pm 0.00012$ \\
\hline
MNIST & SWF barycenter & DIGIT & 0.1 & -- & $\mathbf{904.92 \pm 11.74}$ & $\mathbf{0.0048 \pm 0.00012}$ \\
\hline
MNIST & Liouville PDE-based & DIGIT & 1.0 & 64 & $\mathbf{904.50 \pm 10.87}$ & $\mathbf{0.0075 \pm 0.00017}$ \\
\hline
MNIST & SWF barycenter & DIGIT & 1.0 & -- & $908.42 \pm 11.03$ & $0.0077 \pm 0.00020$ \\
\hline
\end{tabular}
\label{image_MSE_KS}
\end{table}

\begin{table}[!htbp]
\caption{Comparison of accuracy (MSE) and fairness (KS) between SWF barycenter and Liouville PDE-based SWF across Crime and Health datasets for each sensitive attribute. Boldface indicates the lower MSE or KS in mean values.}
\label{accuracy_fairness}
\small
\centering
\begin{tabular}{|c|c|c|c|c|c|c|}
\hline
\textbf{Data} & \textbf{Baseline} & \textbf{Sensitive} & $\mathbf{\lambda}$ & \textbf{Neural ODE} & \textbf{MSE} & \textbf{KS} \\
&  & \textbf{Attributes} &  & \textbf{Layer Size} &  & \\
\hline
Crime & Liouville PDE-based & PctRace & 0.0 & 32 & $\mathbf{112.36 \pm 13.68}$ & $0.0099 \pm 0.00045$ \\
\hline
Crime & SWF barycenter & PctRace & 0.0 & -- & $114.28 \pm 13.99$ & $\mathbf{0.0096 \pm 0.00050}$ \\
\hline
Crime & Liouville PDE-based & PctRace & 0.001 & 32 & $\mathbf{113.60 \pm 13.96}$ & $0.0077 \pm 0.00029$ \\
\hline
Crime & SWF barycenter & PctRace & 0.001 & -- & $114.43 \pm 13.94$ & $\mathbf{0.0058 \pm 0.00022}$ \\
\hline
Crime & Liouville PDE-based & PctRace & 0.01 & 32 & $\mathbf{112.08 \pm 13.55}$ & $\mathbf{0.0061 \pm 0.00037}$ \\
\hline
Crime & SWF barycenter & PctRace & 0.01 & -- & $114.00 \pm 13.80$ & $0.010 \pm 0.00028$ \\
\hline
Crime & Liouville PDE-based & PctRace & 0.1 & 64 & $\mathbf{114.73 \pm 12.91}$ & $0.0084 \pm 0.00043$ \\
\hline
Crime & SWF barycenter & PctRace & 0.1 & -- & $115.20 \pm 13.16$ & $\mathbf{0.0076 \pm 0.00036}$ \\
\hline
Crime & Liouville PDE-based & PctRace & 1.0 & 64 & $\mathbf{228.67 \pm 90.68}$ & $0.0089 \pm 0.00098$ \\
\hline
Crime & SWF barycenter & PctRace & 1.0 & -- & $234.06 \pm 93.12$ & $\mathbf{0.0088 \pm 0.0014}$ \\
\hline
Health & Liouville PDE-based & AGE & 0.0 & 64 & $\mathbf{41.15 \pm 9.61}$ & $0.0079 \pm 0.00067$ \\
\hline
Health & SWF barycenter & AGE & 0.0 & -- & $41.67 \pm 10.04$ & $\mathbf{0.0076 \pm 0.00030}$ \\
\hline
Health & Liouville PDE-based & AGE & 0.001 & 32 & $\mathbf{42.39 \pm 9.64}$ & $\mathbf{0.0068 \pm 0.00020}$ \\
\hline
Health & SWF barycenter & AGE & 0.001 & -- & $42.96 \pm 10.19$ & $0.011 \pm 0.00045$ \\
\hline
Health & Liouville PDE-based & AGE & 0.01 & 32 & $\mathbf{40.86 \pm 9.62}$ & $0.0078 \pm 0.00038$ \\
\hline
Health & SWF barycenter & AGE & 0.01 & -- & $41.31 \pm 9.77$ & $\mathbf{0.0067 \pm 0.00056}$ \\
\hline
Health & Liouville PDE-based & AGE & 0.1 & 64 & $44.00 \pm 8.54$ & $\mathbf{0.0070 \pm 0.00074}$ \\
\hline
Health & SWF barycenter & AGE & 0.1 & -- & $\mathbf{43.66 \pm 8.75}$ & $0.011 \pm 0.00080$ \\
\hline
Health & Liouville PDE-based & AGE & 1.0 & 128 & $265.88 \pm 190.67$ & $\mathbf{0.0050 \pm 0.00051}$ \\
\hline
Health & SWF barycenter & AGE & 1.0 & -- & $\mathbf{265.54 \pm 191.07}$ & $0.0070 \pm 0.0015$ \\
\hline
Health & Liouville PDE-based & sexF & 0.0 & 128 & $41.91 \pm 10.03$ & $\mathbf{0.0093 \pm 0.00069}$ \\
\hline
Health & SWF barycenter & sexF & 0.0 & -- & $\mathbf{41.79 \pm 9.84}$ & $0.0098 \pm 0.00043$ \\
\hline
Health & Liouville PDE-based & sexF & 0.001 & 128 & $\mathbf{41.46 \pm 9.74}$ & $\mathbf{0.0069 \pm 0.00032}$ \\
\hline
Health & SWF barycenter & sexF & 0.001 & -- & $41.99 \pm 9.68$ & $0.0088 \pm 0.00048$ \\
\hline
Health & Liouville PDE-based & sexF & 0.01 & 32 & $41.72 \pm 9.95$ & $\mathbf{0.0068 \pm 0.00039}$ \\
\hline
Health & SWF barycenter & sexF & 0.01 & -- & $\mathbf{40.73 \pm 9.68}$ & $0.0071 \pm 0.00074$ \\
\hline
Health & Liouville PDE-based & sexF & 0.1 & 64 & $\mathbf{43.23 \pm 8.32}$ & $\mathbf{0.0073 \pm 0.00097}$ \\
\hline
Health & SWF barycenter & sexF & 0.1 & -- & $43.77 \pm 8.59$ & $0.0083 \pm 0.00043$ \\
\hline
Health & Liouville PDE-based & sexF & 1.0 & 32 & $\mathbf{265.38 \pm 191.87}$ & $0.012 \pm 0.0024$ \\
\hline
Health & SWF barycenter & sexF & 1.0 & -- & $267.80 \pm 192.02$ & $\mathbf{0.011 \pm 0.0012}$ \\
\hline
Health & Liouville PDE-based & totpay & 0.0 & 32 & $42.09 \pm 9.71$ & $\mathbf{0.0082 \pm 0.0010}$ \\
\hline
Health & SWF barycenter & totpay & 0.0 & -- & $\mathbf{40.43 \pm 9.38}$ & $0.0091 \pm 0.00036$ \\
\hline
Health & Liouville PDE-based & totpay & 0.001 & 32 & $\mathbf{41.49 \pm 9.67}$ & $\mathbf{0.0089 \pm 0.00053}$ \\
\hline
Health & SWF barycenter & totpay & 0.001 & -- & $41.80 \pm 9.82$ & $0.012 \pm 0.00046$ \\
\hline
Health & Liouville PDE-based & totpay & 0.01 & 64 & $\mathbf{41.41 \pm 10.18}$ & $\mathbf{0.0058 \pm 0.00079}$ \\
\hline
Health & SWF barycenter & totpay & 0.01 & -- & $42.36 \pm 10.08$ & $0.010 \pm 0.00047$ \\
\hline
Health & Liouville PDE-based & totpay & 0.1 & 32 & $43.60 \pm 8.41$ & $0.0084 \pm 0.00049$ \\
\hline
Health & SWF barycenter & totpay & 0.1 & -- & $\mathbf{43.56 \pm 8.01}$ & $\mathbf{0.0060 \pm 0.00074}$ \\
\hline
Health & Liouville PDE-based & totpay & 1.0 & 32 & $265.81 \pm 191.23$ & $\mathbf{0.0097 \pm 0.00090}$ \\
\hline
Health & SWF barycenter & totpay & 1.0 & -- & $\mathbf{264.76 \pm 190.79}$ & $0.013 \pm 0.0019$ \\
\hline
\end{tabular}
\label{crime_health_MSE_KS}
\end{table}


%
\begin{figure}[hbt]
    \centering
    \includegraphics[width=0.9 \linewidth]{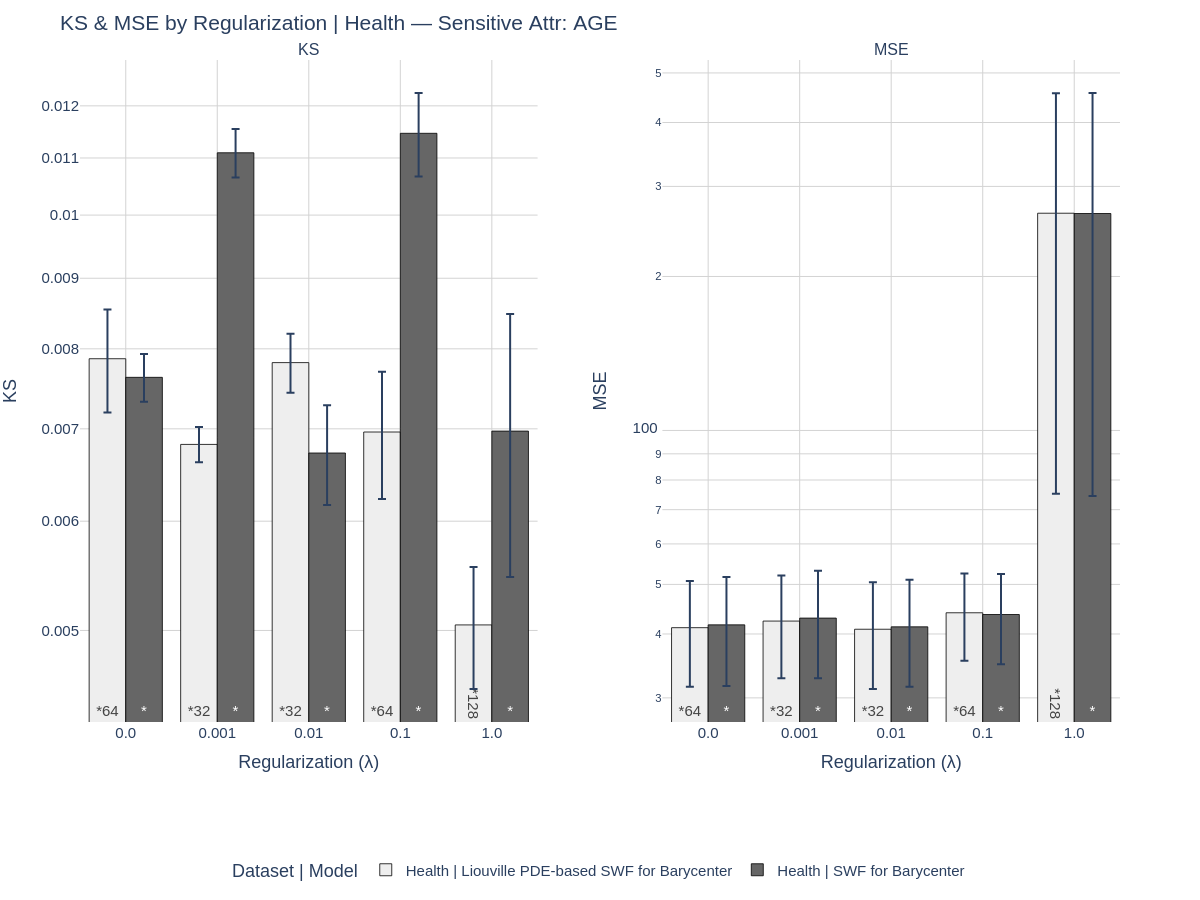}
    \caption{Health Spending dataset on \textit{AGE}. KS (left) and MSE (right) as functions of the regularization parameter $\lambda$ under sensitive-attribute conditioning on \textit{AGE}. Results compare the Liouville PDE-based SWF barycenter and the vanilla SWF barycenter. Error bars indicate variability across runs; numbers inside bars denote the neural ODE layer width.
\label{health_care_age_reg}
}
\end{figure}
\begin{figure}[hbt]
    \centering
    \includegraphics[width=0.9 \linewidth]{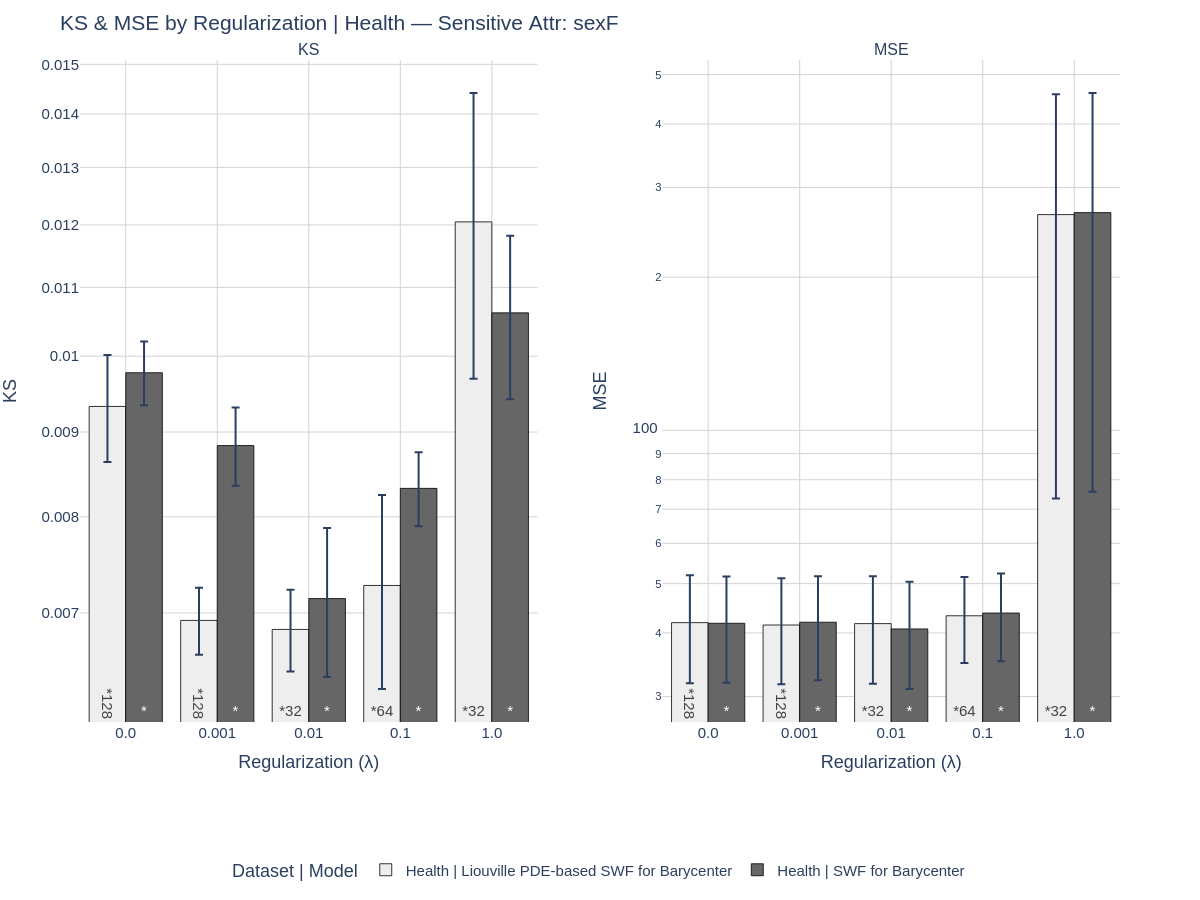}
    \caption{Health Spending dataset on \textit{sexF}. KS (left) and MSE (right) as functions of the regularization parameter $\lambda$ under sensitive-attribute conditioning on \textit{sexF}. Results compare the Liouville PDE-based SWF barycenter and the vanilla SWF barycenter. Error bars indicate variability across runs; numbers inside bars denote the neural ODE layer width. 
 \label{health_care_sexF_reg}
}
\end{figure}

\subsection{Computational usage}
We additionally assess the computational overhead introduced by the Liouville PDE. To isolate the implementation effects, we compare the wall-clock execution time and peak memory usage between the vanilla SWF and its Liouville PDE under variations in ODE layer size and regularization. Across both datasets, the PDE constraint produces only marginal increases in runtime and memory consumption, indicating that the fairness and accuracy improvements reported earlier are obtained at negligible computational cost.
%

\begin{figure}[hbt]
    \centering
    \includegraphics[width=0.95\linewidth]{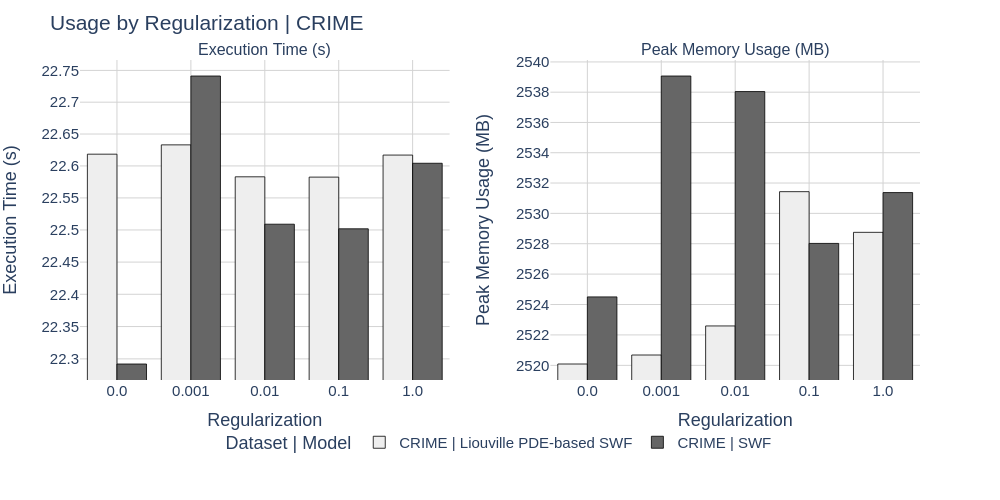}
\caption{Crime dataset. Execution time (left) and peak memory usage (right) as functions of the regularization parameter $\lambda$ for the Liouville PDE-based SWF and the vanilla SWF baseline. Across the practical range of regularization values, execution time remains nearly constant and peak memory usage varies only modestly, indicating that PDE-based regularization introduces negligible computational overhead.} 
\label{crime_execution_time_lambda}
\end{figure}

\begin{figure}[hbt]
    \centering
    \includegraphics[width=0.95\linewidth]{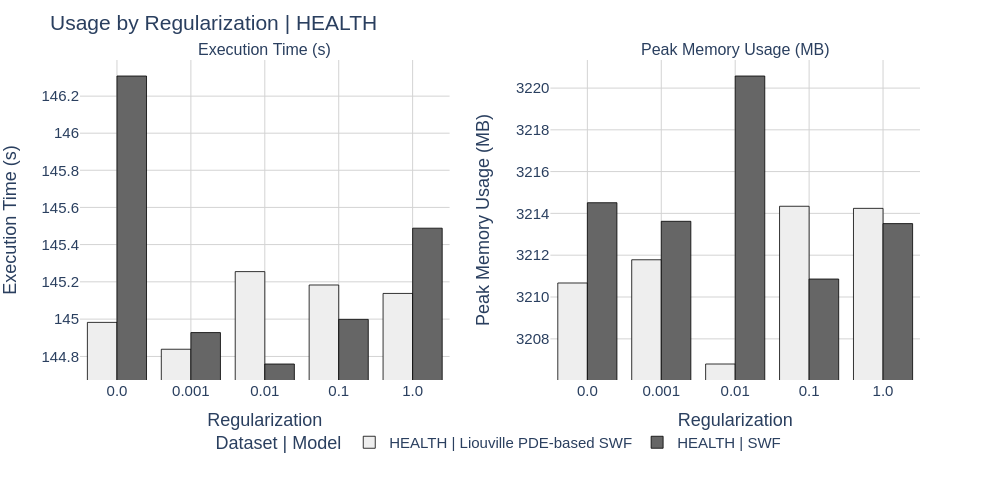}
\caption{Health Spending dataset. Execution time (left) and peak memory usage (right) as functions of the regularization parameter $\lambda$ for the Liouville PDE-based SWF and the vanilla SWF baseline. Across the practical range of $\lambda$, execution time and memory usage remain stable, with only minor increases at the largest regularization level, confirming the computational scalability of the PDE-based formulation.} 
\label{healthcare_execution_time_lambda}
\end{figure}

\begin{figure}[hbt]
    \centering
    \includegraphics[width=0.95\linewidth]{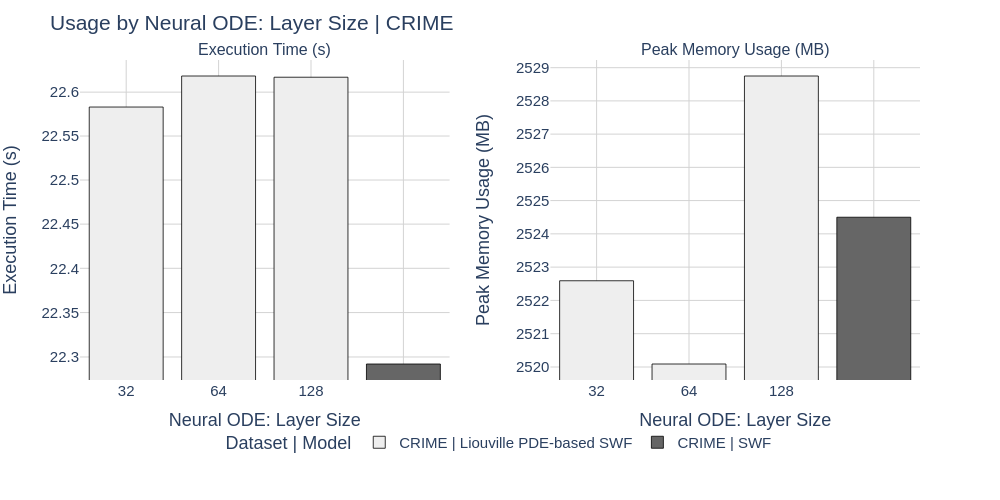}
    \caption{Crime dataset. Execution time (left) and peak memory usage (right) as functions of the neural ODE layer width for the Liouville PDE-based SWF; the vanilla SWF baseline (black bars) does not employ a neural ODE and is therefore shown without an associated layer width. Across configurations, runtime differences remain below $0.2\,\mathrm{s}$ and peak memory usage varies by less than $10\,\mathrm{MB}$, indicating that the PDE-based regularization introduces negligible computational overhead.} 
 \label{crime_execution_time_neural_ODE_layer_width}
\end{figure}

\begin{figure}[hbt]
    \centering
    \includegraphics[width=0.95\linewidth]{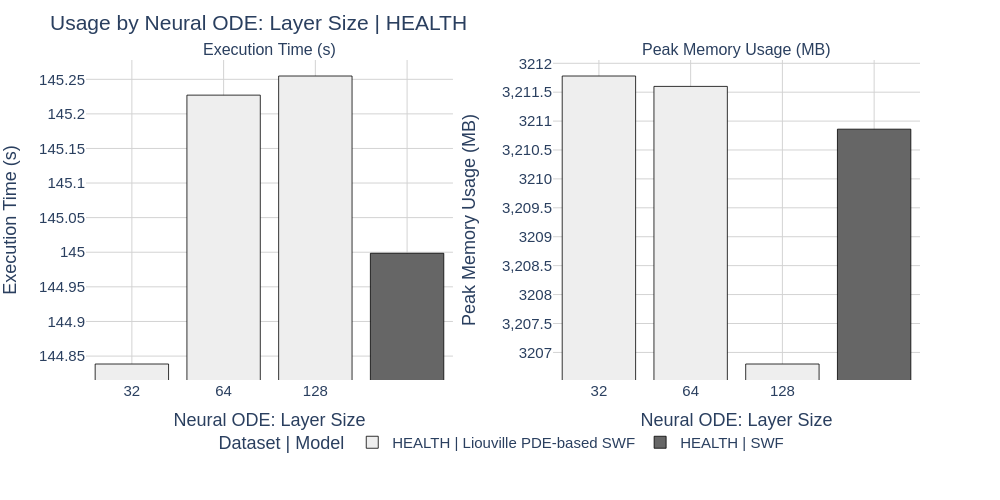}
\caption{Execution time (left) and peak memory usage (right) as functions of the neural ODE layer width for the Liouville PDE-based SWF; the vanilla SWF baseline (black bars) does not employ a neural ODE and is therefore shown without an associated layer width. Across architectures, runtime differences remain modest (typically below $1\,\mathrm{s}$) and peak memory usage varies by approximately $5$--$10\,\mathrm{MB}$, indicating that the PDE-based regularization introduces minimal computational overhead.} 
 \label{healthcare_execution_time_neural_ODE_layer_width}
\end{figure}

Across variations in neural ODE layer width and regularization strength, both the Liouville PDE–based SWF and the vanilla SWF exhibit stable and nearly invariant computational cost. For the Communities and Crime dataset, execution time varies within a narrow range (approximately 17.1 - 18.2 s), while peak memory usage remains tightly bounded (about 2477 - 2498 MB) (Table \ref{crime_execution_time_lambda}, \ref{crime_execution_time_neural_ODE_layer_width}). Similarly, for the Health Care Spending Cost dataset, execution times fluctuate modestly with architecture and regularization (approximately 174 - 180 s), and peak memory usage remains confined to a narrow band (about 3358 - 3369 MB) (Table \ref{healthcare_execution_time_lambda}, \ref{healthcare_execution_time_neural_ODE_layer_width}). The vanilla SWF baseline, which does not employ a neural ODE, exhibits comparable runtime and memory profiles. These small variations indicate that the observed improvements in fairness (KS) and accuracy (MSE) are not attributable to increased computational load, but instead reflect genuine modeling differences introduced by the Liouville PDE-based formulation.



\end{document}